\documentclass[lettersize,journal]{IEEEtran}
\usepackage{amsmath,amsfonts}
\usepackage{algorithmic}
\usepackage{algorithm}
\usepackage{array}
\usepackage[caption=false,font=normalsize,labelfont=sf,textfont=sf]{subfig}
\usepackage{textcomp}
\usepackage{stfloats}
\usepackage{url}
\usepackage{verbatim}
\usepackage{graphicx}

\usepackage{bm}
\usepackage{xcolor}

\usepackage{cite}          
\usepackage[hidelinks]{hyperref}  
\usepackage{placeins}
\usepackage{mathtools}             
\allowdisplaybreaks[1]             
\usepackage{tabularx}
\usepackage{booktabs}
\usepackage{makecell}
\usepackage{pifont} 

\usepackage{graphicx}
\usepackage{subfig}
\usepackage{multirow}
\usepackage{subfig}

\usepackage{hyperref}

\hypersetup{
  colorlinks=true,
  citecolor=blue,
  linkcolor=blue,
  urlcolor=black
}

\usepackage{amsthm}

\theoremstyle{remark}
\newtheorem{remark}{Remark}

\begin{document}

\title{RoomEditor++: A Parameter-Sharing Diffusion Architecture for High-Fidelity Furniture Synthesis}

\author{Qilong Wang, Xiaofan Ming, Zhenyi Lin, Jinwen Li, Dongwei Ren, Wangmeng Zuo, Qinghua Hu

\thanks{(Qilong Wang and Xiaofan Ming contributed equally to this work.) (Corresponding author: Dongwei Ren.)}
\thanks{Qilong Wang, Xiaofan Ming, Zhenyi Lin, Jinwen Li, Dongwei Ren are with the School of Artificial Intelligence, Tianjin University, Tianjin 300350, China (e-mail: qlwang@tju.edu.cn; xiaofanming@tju.edu.cn; linzhenyi@tju.edu.cn; lijinwen@tju.edu.cn; rendw@tju.edu.cn).}
\thanks{Wangmeng Zuo is with the School of Computer Science and Technology, Harbin Institute of Technology, Harbin 150001, China (e-mail: cswmzuo@gmail.com).}
\thanks{Qinghua Hu is with the School of Artificial Intelligence, Tianjin University, Tianjin 300350, China, and also with the Engineering Research Center of City Intelligence and Digital Governance, Ministry of Education, Tianjin 300350, China (e-mail: huqinghua@tju.edu.cn)}
}



\maketitle

\begin{abstract}

Virtual furniture synthesis, which seamlessly integrates reference objects into indoor scenes while maintaining geometric coherence and visual realism, holds substantial promise for home design and e-commerce applications. 
However, this field remains underexplored due to the scarcity of reproducible benchmarks and the limitations of existing image composition methods in achieving high-fidelity furniture synthesis while preserving background integrity. 
To overcome these challenges, we first present RoomBench++, a comprehensive and publicly available benchmark dataset tailored for this task. It consists of 112,851 training pairs and 1,832 testing pairs drawn from both real-world indoor videos and realistic home design renderings, thereby supporting robust training and evaluation under practical conditions. 
Then, we propose RoomEditor++, a versatile diffusion-based architecture featuring a parameter-sharing dual diffusion backbone, which is compatible with both U-Net and DiT architectures. This design unifies the feature extraction and inpainting processes for reference and background images. Our in-depth analysis reveals that the parameter-sharing mechanism enforces aligned feature representations, facilitating precise geometric transformations, texture preservation, and seamless integration. Extensive experiments validate that RoomEditor++ is superior over state-of-the-art approaches in terms of quantitative metrics, qualitative assessments, and human preference studies, while highlighting its strong generalization to unseen indoor scenes and general scenes without task-specific fine-tuning. The dataset and source code are available at \url{https://github.com/stonecutter-21/roomeditor}.
\end{abstract}

\begin{IEEEkeywords}
Image synthesis, diffusion model, home design.
\end{IEEEkeywords}

\IEEEpubidadjcol
\section{Introduction}
\begin{figure*}[h]
    \centering
    \includegraphics[width=\textwidth]{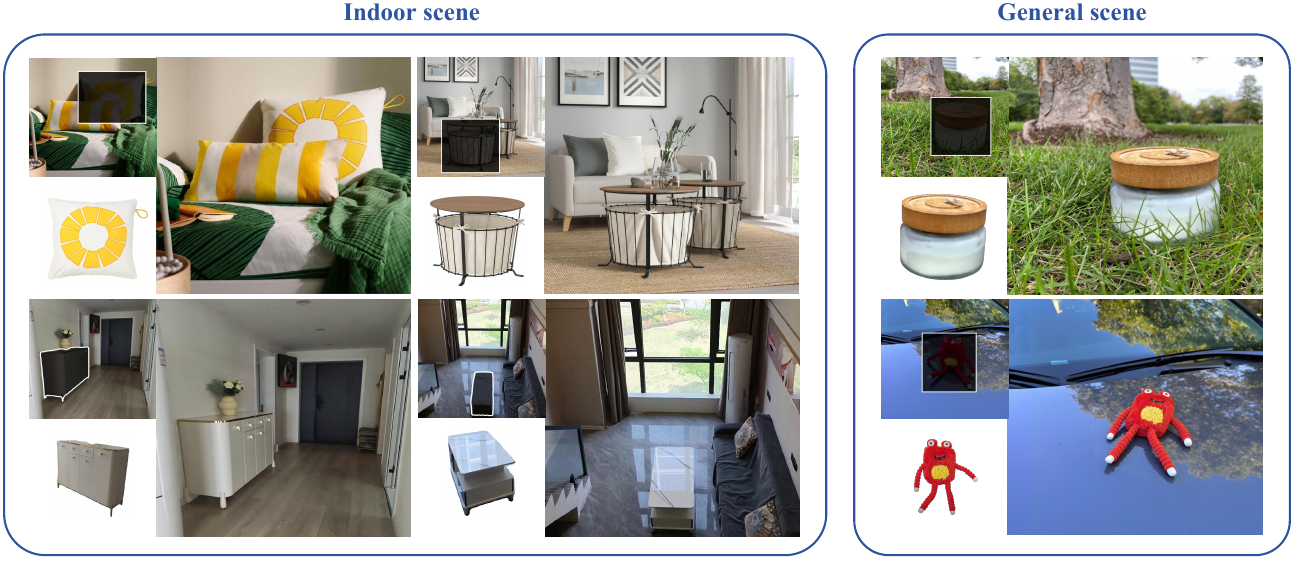}
    \caption{Furniture synthesis with our RoomEditor++ integrates reference objects into environments with geometric coherence and visual fidelity. Moreover, RoomEditor++ exhibits remarkable generalization capabilities across a wide range of unseen scenes and objects without task-specific fine-tuning.}
    \label{fig:results}
\end{figure*}

\IEEEPARstart{R}{ecent} advances in augmented reality and computer vision have revolutionized virtual product visualization for e-commerce, with profound implications for indoor scene synthesis. Within the rapidly expanding home design market (expected to reach around 250 billion dollars in next decade \cite{mrfr2025}), intelligent image synthesis systems have emerged as critical tools for consumer decision-making. These systems enable in-situ furniture visualization by digitally integrating selected items into user-provided room images, as shown in \autoref{fig:results}, offering significant potential to redefine design workflows and retail experiences.
Despite progress in generic image composition, virtual furniture synthesis remains underdeveloped due to two key limitations: the scarcity of ready-to-use benchmark datasets, and the inadequacy of existing image composition methods in achieving high-fidelity furniture synthesis while guaranteeing background integrity. 

First, for the issues of available benchmark, Amazon’s dataset \cite{diffuse_to_choose} highlights the value of domain-specific data, but its restricted access stifles community-wide research. 
Existing 3D-centric resources, e.g., InteriorNet \cite{Interiornet}, 3D-FUTURE \cite{3d-future}, and 3D-FRONT \cite{3d-front}, provide accessible 3D indoor models, but constructing reasonable reference–background pairs demands meticulous viewpoint selection and scene setup, creating significant barriers for data acquisition. More critically, 3D-derived samples are inherently synthetic, lacking real-world traits like natural variability, texture noise, etc. This gap limits the development of generalizable methods in practical scenarios.   
To address this data bottleneck, we present RoomBench++, a comprehensive and publicly accessible benchmark designed specifically for furniture synthesis under practical conditions. RoomBench++ comprises 112,851 training pairs and 1,832 testing pairs, including two complementary subsets: a realistic-scene subset sourced from professional home design renderings~\cite{lin2025roomeditor}, and a real-scene subset derived from heterogeneous real-world indoor video sequences.
This benchmark can be readily used to act a foundation for training and evaluation furniture synthesis methods under practical conditions. 

Second, the challenge arises from technical limitations of existing image composition methods~\cite{pbe,anydoor,mimicbrush,song2025insert}, 
which generally fail to satisfy the visual fidelity requirements by practical deployment in home design. Even minor discrepancies in geometric alignment or textural congruence can compromise perceptual immersion and erode user trust. Encoder-based methods like AnyDoor~\cite{anydoor}, which extract reference features using pre-trained models, e.g., CLIP~\cite{clip}, for injection into diffusion backbones, frequently overlook fine-grained structural details essential for complex furniture integration.
Dual-branch U-Net architectures such as MimicBrush~\cite{mimicbrush} exacerbate this issue by processing reference and background images independently, leading to feature misalignments that manifest as visual artifacts or distorted perspectives. 
Recently, DiT-based approaches~\cite{song2025insert,huang2025dreamfuse} still suffer from latent space inconsistencies due to the segregated processing pipeline, potentially limiting performance of real-world scene synthesis. The corresponding visualization results can be found in \autoref{sec:exp}. Collectively, these limitations undermine synthesis coherence and practical utility.

To address the technical challenges, we propose RoomEditor++, a versatile diffusion-based architecture, which employs a parameter-sharing dual diffusion backbone, and is compatible with both U-Net \cite{ronneberger2015u} and DiT \cite{Peebles2022DiT} architectures. This design unifies the feature extraction and inpainting processes for reference and background images, intrinsically enforcing aligned feature representations.
Our analysis reveals that such parameter-sharing mechanism is critical for facilitating precise geometric transformations, texture preservation, and seamless integration, effectively addressing the core limitations of prior methods.
Quantitative and qualitative evaluations on RoomBench++ confirm that our RoomEditor++ outperforms existing state-of-the-art methods across objective metrics, visual inspections, and human perceptual evaluations. Furthermore, our RoomEditor++ demonstrates promising generalization to unseen indoor configurations (e.g., samples from 3D-FUTURE \cite{3d-future}), and extrapolates effectively to diverse scenes and object categories (e.g., samples from DreamBooth \cite{dreambooth}) without task-specific fine-tuning, as illustrated in~\autoref{fig:results}, also validating the robustness of its unified design. 
The contributions of this work are summarized as follows:
\begin{itemize}
    \item 
    We construct RoomBench++, an open-source benchmark curated for home design, which merges real-scene and realistic-scene data including 112,851 training pairs and 1,832 testing pairs. These samples are either derived from real-world indoor scenarios or professional home design renderings, fully aligning with the practical requirements of furniture synthesis tasks. 

    \item We propose RoomEditor++, a concise architecture for high-fidelity furniture synthesis, which adopts a parameter-sharing dual diffusion backbone, and is compatible with U-Net and DiT architectures. 
    We further provide a comprehensive analysis for this design from the perspective of feature consistency. 
    
    \item Extensive experiments validate that our RoomEditor++ achieves state-of-the-art performance in quantitative, qualitative evaluations, and human perception studies for home design. It also demonstrates remarkable generalization across diverse scenes and objects without fine-tuning.
\end{itemize}

This work was previously presented as a conference paper~\cite{lin2025roomeditor}, involving a RoomBench dataset and a RoomEditor method. The current paper introduces three major extensions. 
(\emph{i}) RoomBench++ extends the benchmark by introducing a much more large-scale real-world subset (more than 100,000 image pairs) to enable more robust evaluation and training under practical conditions, while the original RoomBench \cite{lin2025roomeditor} only includes realistic home design renderings (about 7,000 image pairs). Moreover, data processing pipeline is designed in an almost fully automated manner. 
(\emph{ii}) From technical perspective, RoomEditor++ is a versatile framework compatible with both U-Net and DiT backbones to enhance the ability of high-fidelity furniture synthesis, while previous RoomEditor~\cite{lin2025roomeditor} only supports U-Net backbone. 
(\emph{iii}) More experiments are conducted on multiple benchmarks, and the results show that our RoomEditor++ improves previous RoomEditor by a large margin across diverse evaluation metrics.

The remainder of this paper is organized as follows: \autoref{Sec:related_work} presents a comprehensive review of related work; \autoref{sec:data_overreview} details the RoomBench++ dataset; \autoref{sec:method} describes the RoomEditor++ architecture and provides in-depth analysis;  \autoref{sec:exp} reports extensive experimental validations, and finally \autoref{sec:conclusion} concludes our work.

\captionsetup[subfloat]{font=scriptsize}


\section{RELATED WORK}
\label{Sec:related_work}
\subsection{Furniture Synthesis for Home Design}
Most relevant research with furniture synthesis is virtual try-on~\cite{xie2023gp,viton, viton_hd, tryongan, stableviton, ootdiffusion, mmtryon,catvton,he2024wildvidfit,chong2025catv2ton}, which has primarily focused on fashion applications, emphasizing fabric textures, body pose estimation, and occlusion handling.
While several virtual try-on methods~\cite{stableviton, ootdiffusion, mmtryon} employ dual U-Net architectures and require to model garment–body interactions, whose objectives and challenges differ from those in furniture synthesis. In particular, furniture synthesis requires handling diverse and rigid object shapes, whereas clothing in virtual try-on typically exhibits more consistent forms, leading to different integration challenges. Meanwhile, indoor scene synthesis specifically designed for furniture placement remains underexplored. A few studies~\cite{chen2023generating, roomdiffusion} have explored interior design through text-to-image generation but provide limited control over individual furniture placement. Diffuse-to-Choose~\cite{diffuse_to_choose} addresses furniture synthesis by incorporating reference features via a U-Net encoder augmented with FiLM~\cite{film} layers on a large-scale indoor dataset. 
However, its training and testing sets are proprietary, limiting reproducibility and further research. Although HomeDiffusion~\cite{li2025homediffusion} explores furniture synthesis through multi-view object representation learning (MORL) and background-driven object customization learning (BOCL), its training on synthetic data from 3D-FRONT~\cite{3d-front} still leads to a noticeable gap from real-world scenarios. To bridge these gaps, this paper introduces a ready-to-use, open-source  RoomBench++ benchmark for furniture synthesis, which includes both real-scene and realistically rendered scene data.

\subsection{Image Composition}
Early image composition methods~\cite{drag_and_drop,color_harmonization,sunkavalli2010multi,tao2013error,tsai2017deep,cun2020improving,dovenet} primarily focused on the task of image harmonization, ensuring seamless integration of the foreground with the background. These methods typically relied on manually designed pipelines. In contrast, text-to-image diffusion models~\cite{stable_diffusion,ramesh2022hierarchical,vq_diffusion,photorealistic} have enabled the automatic integration of specific objects into diverse contexts while preserving their identity and attributes. By extracting pseudo-words, Textual Inversion \cite{textual_inversion}, DreamBooth \cite{dreambooth}, and methods \cite{avrahami2022blended, hertz2022prompt, kawar2023imagic, kim2022diffusionclip, liu2023more, nichol2021glide, dreamartist, blip_diffusion, custom_diffusion} rely on fine-tuning with textual prompts, but often produce unstable backgrounds. Semantic image composition methods, such as DreamPaint~\cite{dreamPaint}, PBE~\cite{pbe}, CustomNet~\cite{customnet}, and ControlCom~\cite{controlcom}, which focus to insert objects into predefined scenes and often struggle to preserve fine-grained details in complex scenes when using CLIP~\cite{clip} image encoder. AnyDoor~\cite{anydoor} addresses some of these challenges by leveraging ControlNet~\cite{controlnet} and DINOv2~\cite{dinov2} to improve texture fidelity and reference-object consistency. In contrast, MimicBrush~\cite{mimicbrush} employs a dual U-Net architecture~\cite{tunneltryon,reference_only_control,flashface,wearanyway,animate_anyone,magicanimate}, which has proven effective in capturing multi-scale reference features, thereby enabling flexible image editing and generating high-fidelity synthesis.  

Recent advancements, especially in the field of text-to-image diffusion models, have enabled automated and adaptive approaches; however, several challenges still remain. For example, In-context LoRA~\cite{huang2024context} exploits DiT's~\cite{Peebles2022DiT} in-context learning for thematic image generation. Diptych Prompt~\cite{shin2025large}, on the other hand, enables training-free zero-shot reference-based generation via Flux ControlNet. However, both still lack controllable image insertion and rely solely on text-driven contextual reconstruction. In contrast, Insert Anything~\cite{song2025insert} and DreamFuse~\cite{huang2025dreamfuse} extend in-context ability of DiT to support image-based reference inputs. DreamFuse further enhances texture fidelity and reference-object consistency, achieving improved seamless integration, particularly in complex scenes. However, DreamFuse is primarily designed as a generic fusion model rather than being tailored to indoor furniture synthesis, and its training data mainly consists of synthesized fusion images instead of real home-design layouts. Therefore, this paper proposes a RoomEditor++ method to ensure high-fidelity object preservation and context-aware placement, achieving both seamless integration and high fidelity in home design.


\begin{figure}[!t]
    \captionsetup[subfloat]{font=scriptsize}  
    \centering
    \subfloat[Realistic-scene data]{%
        \includegraphics[width=0.22\textwidth]{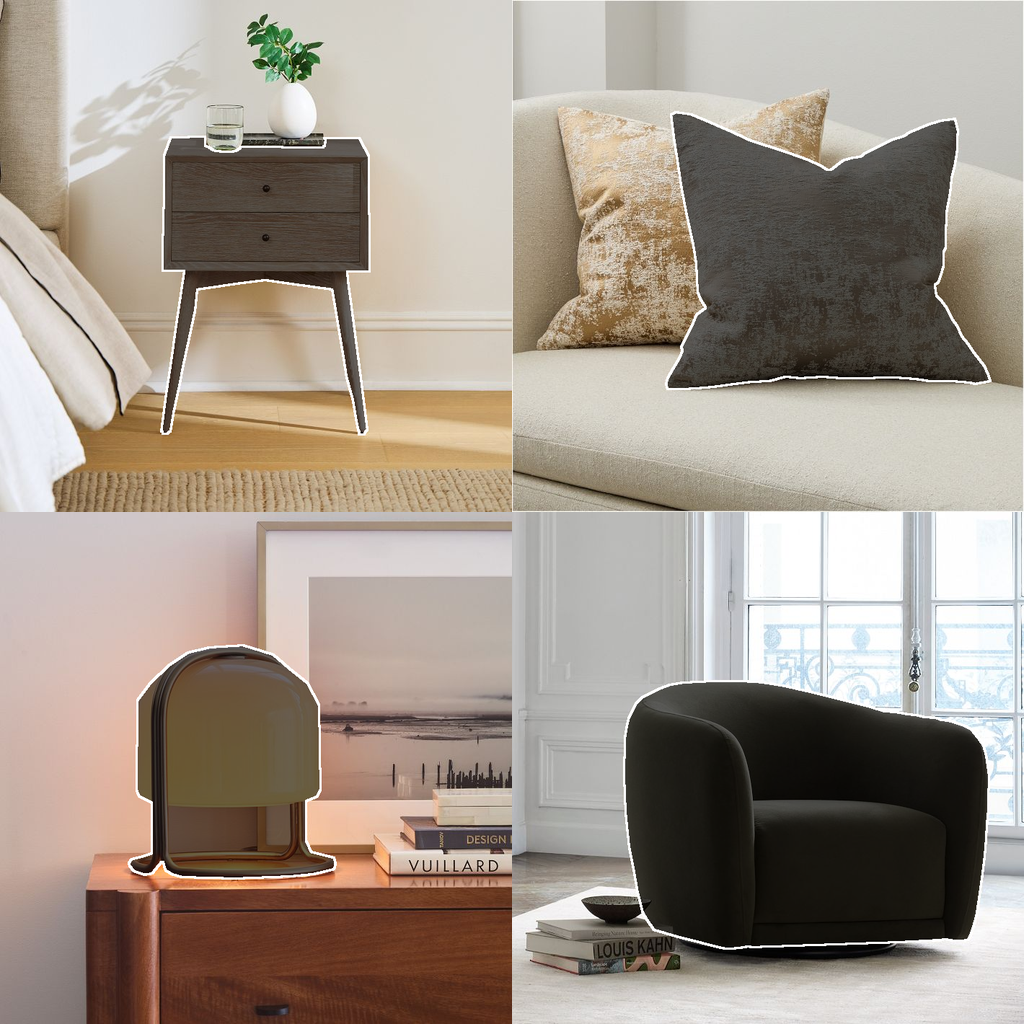}
        \label{fig:realistic_data}
    }
    \hfill
    \subfloat[Real-scene data]{%
        \includegraphics[width=0.22\textwidth]{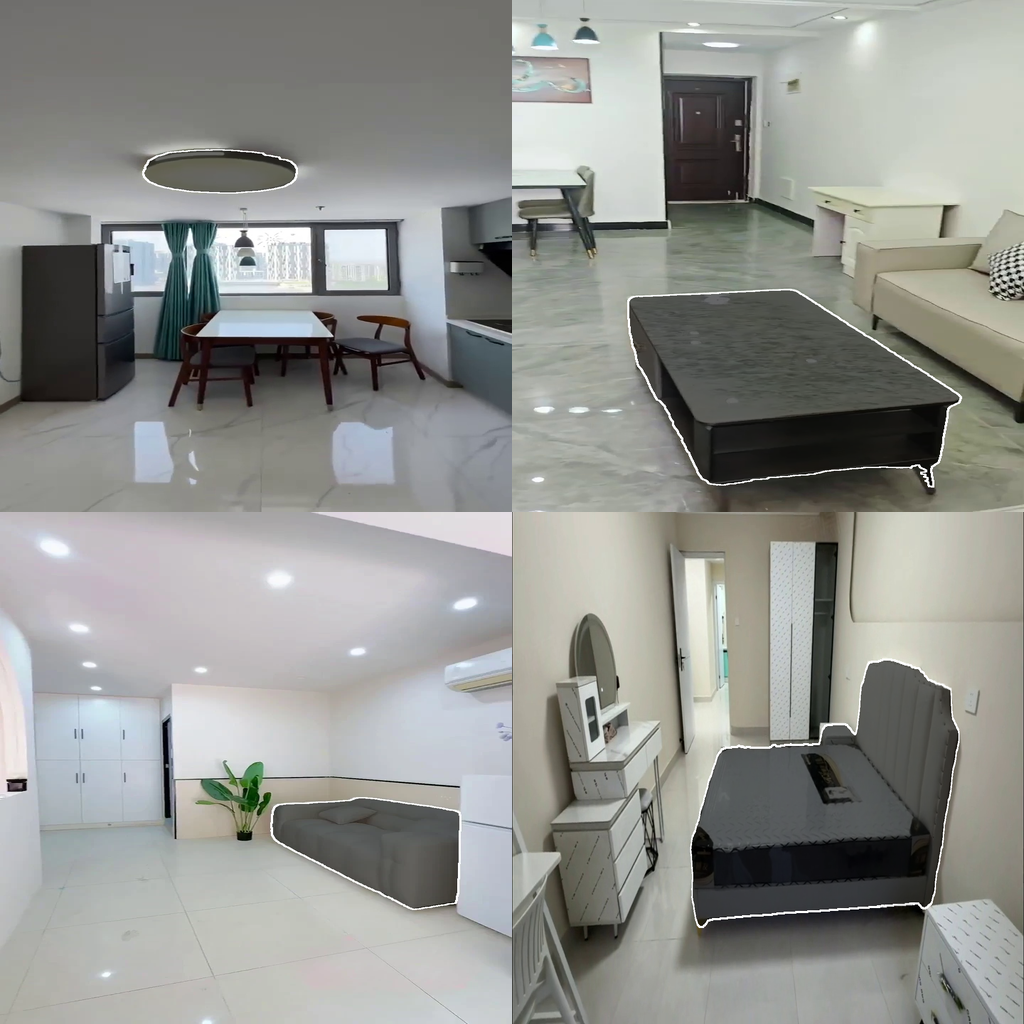}
        \label{fig:real_data}
    }
    \caption{
    The annotations for Roombench++. For the realistic-scene data, the labels are manually annotated. In contrast, for the real-scene data, the annotations are produced using the multimodal segmentation model Sa2Va~\cite{sa2va}.
    }
    \label{fig:main}
\end{figure}
\begin{figure*}[!th]
    \centering

    \subfloat[Diagram of realistic-scene subset construction.]{
        \includegraphics[width=0.70\textwidth]{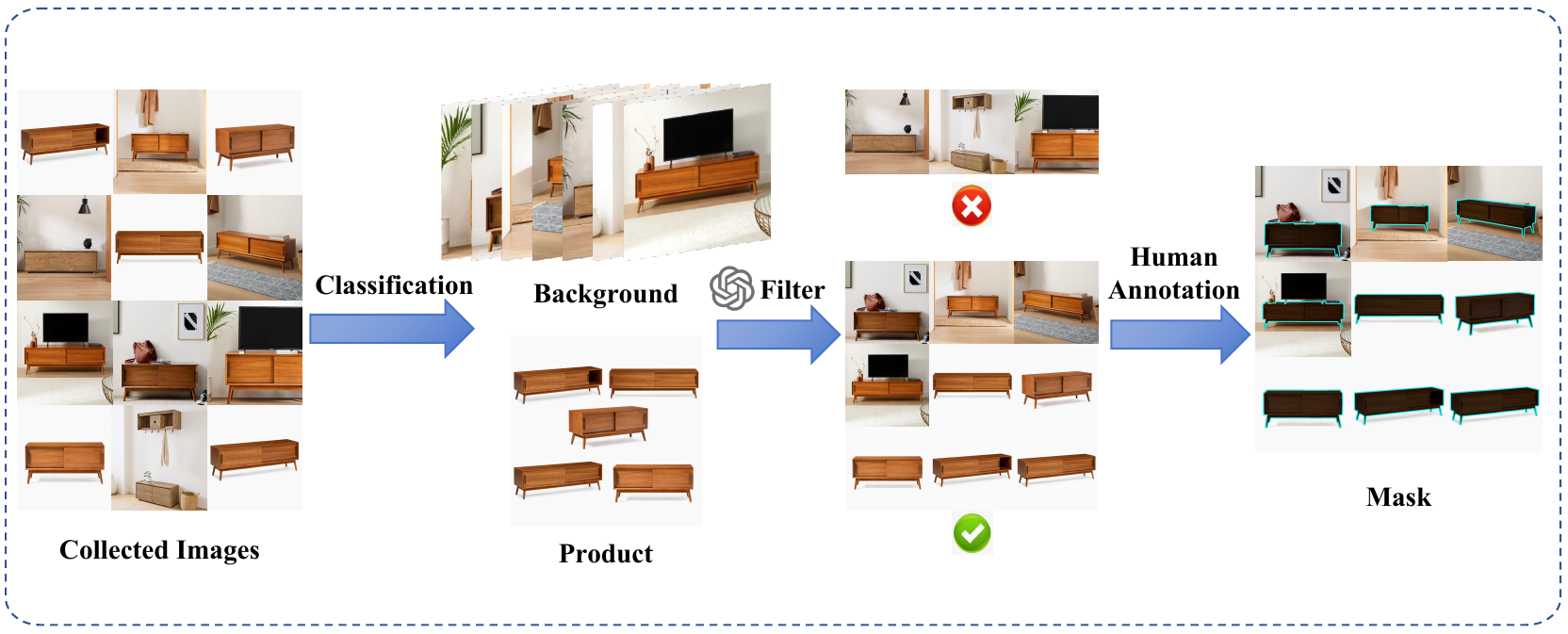}
        \label{fig:realistic_data_construction_a}
    }
    \hfill
    \subfloat[Categories in realistic-scene subset.]{
        \includegraphics[width=0.26\textwidth]{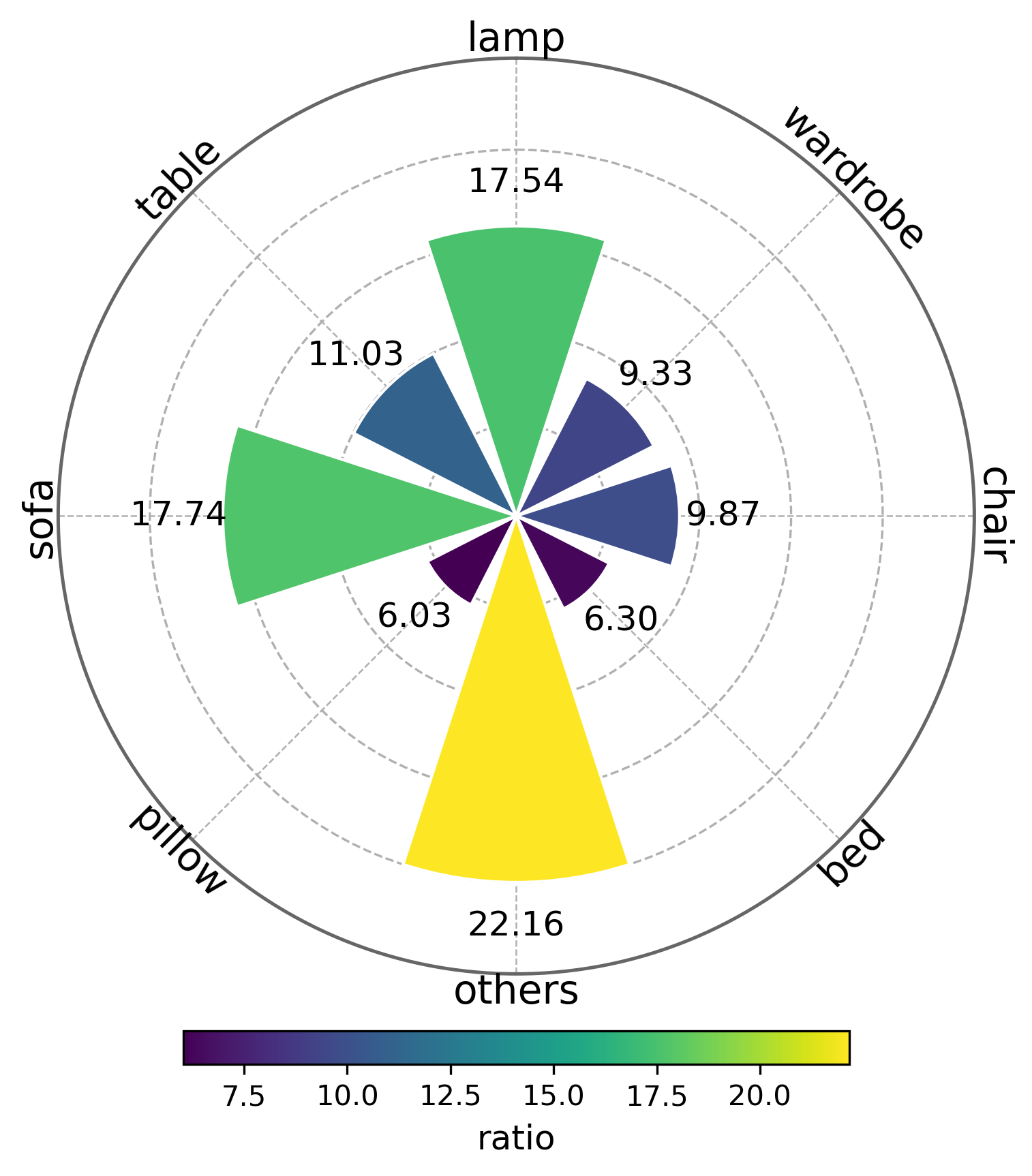}
        \label{fig:realistic_data_construction_b}
    }

    \captionsetup{justification=centering}
    \caption{
        Overview of constructing realistic-scene subset: (a) data construction and (b) furniture categories. 
        As shown in (a), after classifying the images as either product or background images, we employed GPT-4o~\cite{achiam2023gpt} to assist with data filtering.
        (b) shows the statistics of categories in realistic-scene subset.
    }

    \label{fig:realistic_data_construction}
\end{figure*}

\section{RoomBench++ Dataset}
\label{sec:data_overreview}
To address issue of data scarcity in virtual furniture synthesis, we construct an open-source RoomBench++ by extending our previous RoomBench~\cite{lin2025roomeditor}. 
This section details full construction pipeline of RoomBench++, including data collection, rigorous filtering, and annotation strategies, ensuring the utility to train and assess high-fidelity synthesis models.

\subsection{Dataset Overview}
RoomBench++ consists of two complementary subsets with distinct functional orientations: a realistic-scene subset that delivers photorealistic and professionally curated design contexts, and a real-scene subset that captures real-world indoor environments. 
Specifically, realistic-scene data consists of high-quality home design renderings created by professional designers, featuring standardized spatial layouts and details. 
In contrast, real-scene data is directly extracted from real-world indoor video sequences, preserving the natural variability, dynamic lighting, and occluded details inherent to actual living spaces. These dual data sources, with clear definitions and complementary strengths, collectively cover the core demands of diverse furniture synthesis scenarios.
The detailed split and composition of each subset are elaborated as follows. 

\subsubsection{\textbf{Realistic-scene Subset}}
Derived from professional home design renderings (consistent with the original RoomBench \cite{lin2025roomeditor}), this subset provides well-structured and style-coherent ``furniture–background" pairs to support fine-grained synthesis training.
\emph{{Training split}} includes 5,288 foreground furniture images and 4,094 background images. Each furniture image is paired with 1–2 semantically matching background images (e.g., a dining chair with a dining room backdrop), generating 7,298 stylistically consistent training pairs.
\emph{{Testing split}} forms 895 testing pairs that focus on evaluating texture coherence and style alignment, comprising 660 furniture reference images and 895 background images. 

\begin{figure*}[!t]
    \centering

    \subfloat[Diagram of real-scene subset construction.]{
        \includegraphics[width=0.70\textwidth]{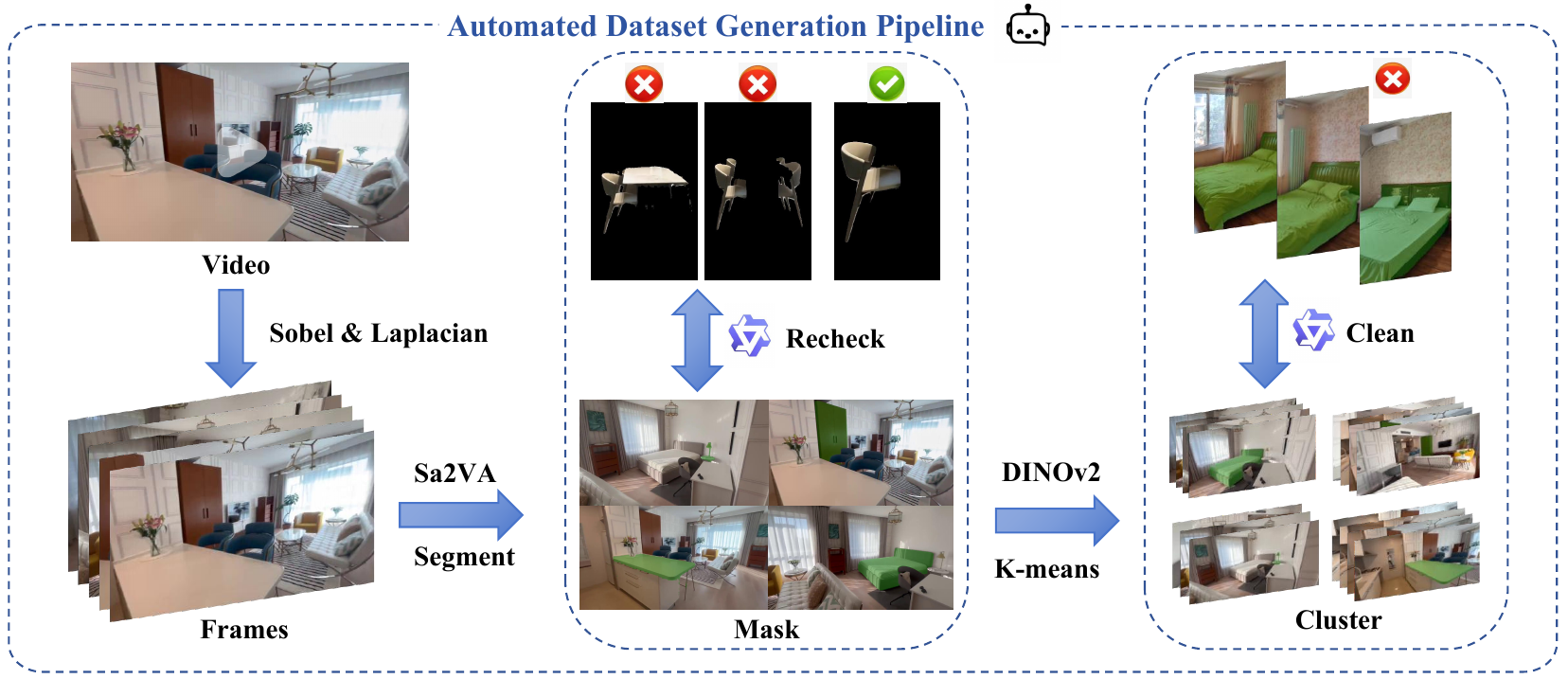}
        \label{fig:data_construction_a}
    }
    \hfill
    \subfloat[Categories in real-scene subset.]{
        \includegraphics[width=0.26\textwidth]{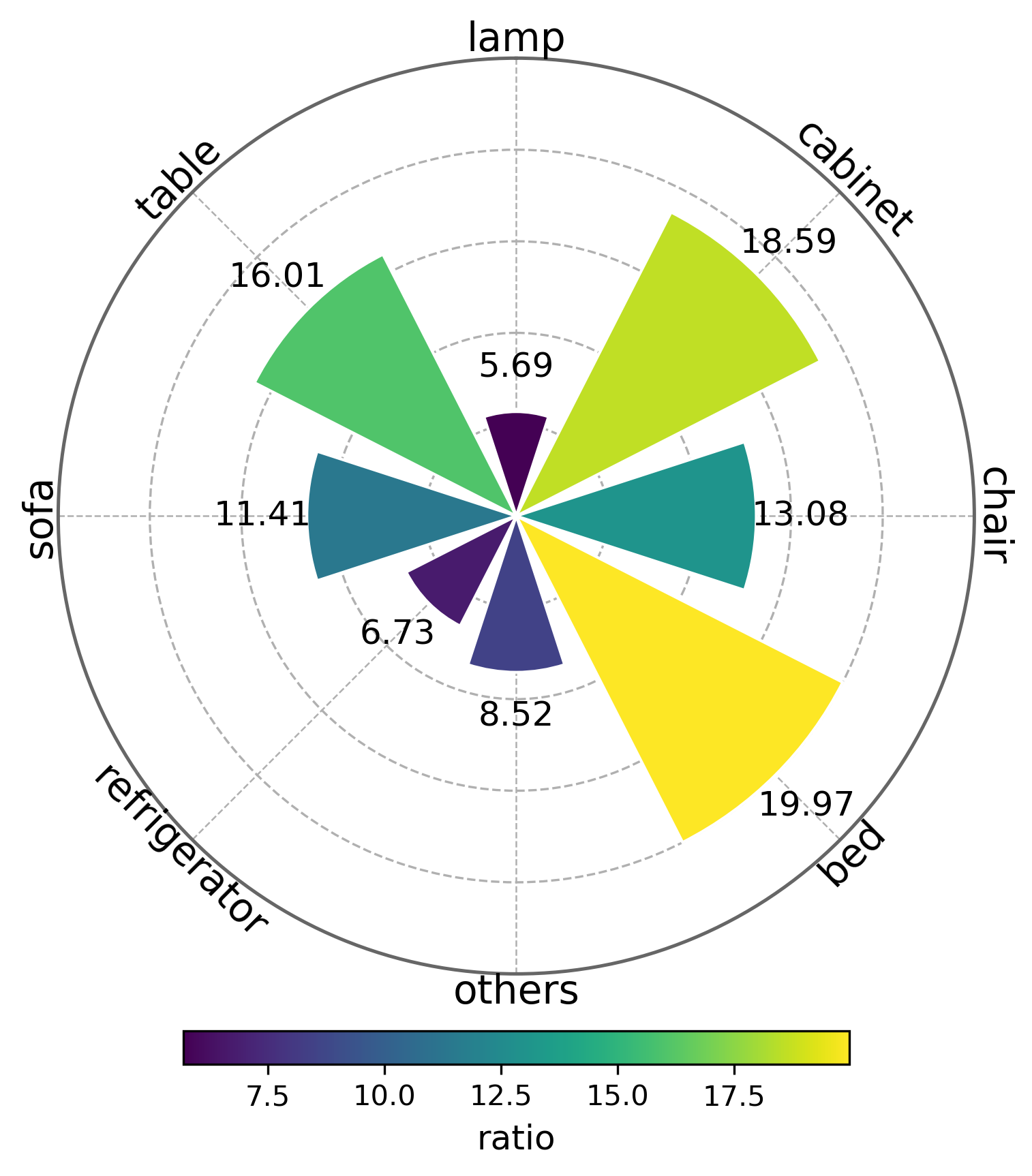}
        \label{fig:data_construction_b}
    }

    \captionsetup{justification=centering}
    \caption{
        Overview of constructing real-scene subset: (a) data construction and (b) furniture categories. 
        In (a), after frame extraction from the video data, it is processed using multimodal large models and traditional machine learning methods, resulting in a nearly fully automated pipeline to obtain the final real-world scene dataset. 
        (b) shows the statistics of categories in real-scene subset.
    }

    \label{fig:data_construction}
\end{figure*}
\subsubsection{\textbf{Real-scene Subset}}
Sourced from heterogeneous real-world indoor video sequences, this subset reflects the natural variability of furniture poses, lighting conditions, and scene layouts in practical scenarios.
For \emph{training split}, we first extract 90,726 frames from raw videos, then identify the same furniture object across different frames to form ``reference–target" pairs (leveraging object consistency across video frames). This process yields 105,553 high-quality training pairs, where each pair shares the same furniture instance but differs in scene context or object pose.
For testing split, we select 1,389 representative frames from independent video clips to evaluate model robustness. For each furniture object, the first frame it appears in is designated as the reference image, while all subsequent frames containing the same object (with natural pose changes) serve as editing targets. As such, 937 testing pairs are generated to access geometric transformation ability.

The key advantages of our RoomBench++ lie in its ready-to-use and practice-oriented nature. Unlike 3D-centric datasets (e.g., 3D-FUTURE \cite{3d-future}, 3D-FRONT \cite{3d-front}) that require laborious viewpoint adjustment and scene reconstruction, our benchmark directly provides annotated image pairs. Moreover, the large-scale real-scene subset (accounting for~94\% of training data) and the professional realistic renderings bridge the gap between model training and practical deployment, significantly enhancing its potential value in promoting furniture synthesis.

\subsection{Data Construction Pipeline}
To ensure data quality while minimizing human labor, we design tailored construction pipelines for the two subsets: a semi-automated workflow for realistic-scene subset for prioritizing annotation precision and a fully automated pipeline for real-scene subset for enabling large-scale collection.

\subsubsection{\textbf{Realistic-scene Subset}}
This subset undergoes a two-stage filtering process to eliminate inconsistent or low-quality pairs, followed by a tiered annotation strategy to simulate both ideal and practical scenarios (see \autoref{fig:realistic_data_construction}{(a)}). 
\autoref{fig:realistic_data_construction}{(b)} presents the furniture category distribution of the realistic-scene training set, covering major home furniture types (e.g., sofas, tables, chairs) to ensure task relevance.

\noindent  
\emph{[Step I] Home Design Renderings Collection}: We meticulously curated the dataset from publicly available webpages on IKEA\footnote{https://www.ikea.com/}, West Elm\footnote{https://www.westelm.com/}, World Market\footnote{https://www.worldmarket.com/}, and other retail websites. 
It is important to emphasize that we only collected publicly accessible content and strictly adhered to the Terms of Service of all source websites. During the data acquisition process, we did not circumvent any access controls or technical protection measures, nor did we obtain any non-public data.

\noindent  
\emph{[Step II] Initial Classification \& AI-Assisted Filtering}: We first manually categorize raw renderings into ``furniture foreground" and ``indoor background" images. We then leverage GPT-4o \cite{achiam2023gpt} to filter invalid pairs, excluding cases where the furniture is partially occluded or not fully visible in the background, and the furniture’s category or color is inconsistent between the reference and background.

\noindent  
\emph{[Step III] Tiered Mask Annotation}: For each valid pair, we annotate pixel-wise furniture masks (see \autoref{fig:main}{(a)}) with varying precision based on the partition. 
For \emph{training split}, masks are manually labeled with sub-pixel precision to ensure high-quality supervision, enabling effective data augmentation (e.g., rotation, scaling) and enhancing model generalization.
For testing split, masks are loosely annotated to mimic the inaccuracies of user-generated labels in practical applications. This setting allows us to evaluate the model’s robustness to annotation noise, a critical metric for real-world deployment.

\subsubsection{\textbf{Real-scene Subset}}
To achieve large-scale collection, we develop an automated construction pipeline, where only testing set requires minimal manual intervention (see \autoref{fig:data_construction}{(a)}). \autoref{fig:data_construction}{(b)} illustrates the statistics of furniture category for real-scene training set, covering nearly all common household furniture types and so ensuring the strong support on the synthesis of daily furniture in various scenarios. The pipeline consists of seven sequential steps. 

\noindent  
\emph{[Step I] Indoor Videos Collection}: We collected data from publicly accessible webpages on platforms such as Beike\footnote{https://bj.ke.com/} and Lianjia\footnote{https://www.lianjia.com/}. The collected videos contain no personal or privacy-sensitive information. We also note that downstream users are required to comply with the original copyright terms of these websites and are prohibited from using the dataset for commercial purposes or in ways that may infringe upon third-party rights. 

\noindent  
\emph{[Step II] Video Decomposition \& Blurry Frames Removal}: We decompose the collected indoor videos into individual frames and compute the Sobel \cite{sobel1968} and Laplacian \cite{marr1980theory} operators to assess the sharpness of each frame. Frames with Sobel gradient variance value below 1600 or Laplacian variance value below 800 are discarded as motion-blurred frames.

\noindent  
\emph{[Step III] Furniture Detection}: Remaining non-blurry frames are fed into a multi-modal large model Qwen2.5-VL \cite{bai2025qwen2} to identify and label furniture categories present in the scenes.

\noindent  
\emph{[Step IV] Pixel-wise Segmentation \& Mask Refinement}: The segmentation model Sa2Va \cite{sa2va} is adopted to generate initial furniture masks. We retain only masks where the largest connected component accounts for $>$95\% of the total mask area, effectively eliminating masks with multiple objects or severe segmentation errors.

\noindent  
\emph{[Step V] Category Verification}: The segmented object is re-verified by Qwen2.5-VL to ensure it matches the target furniture category, filtering out missegmented regions (e.g., confusing a decorative plant with a chair).

\noindent  
\emph{[Step VI] Object Clustering}: For videos containing multiple furniture instances (e.g., a living room with a sofa and two chairs), we extract global features via DINOv2 \cite{dino} and perform clustering to group semantically similar objects. This step ensures that ``reference–target" pairs are formed with the same furniture instance.

\noindent  
\emph{[Step VII] Cleaning \& Testing Set Curation}: Finally, Qwen2.5-VL performs filtering to remove low-quality samples. For the testing set, we manually select pairs with significant pose variations (e.g., a chair viewed from the front vs. side) to better evaluate the model’s geometric transformation capabilities.

\captionsetup[subfloat]{labelformat=empty}

\begin{figure*}[!t] \centering \includegraphics[width=1\textwidth, height=0.33\textheight]{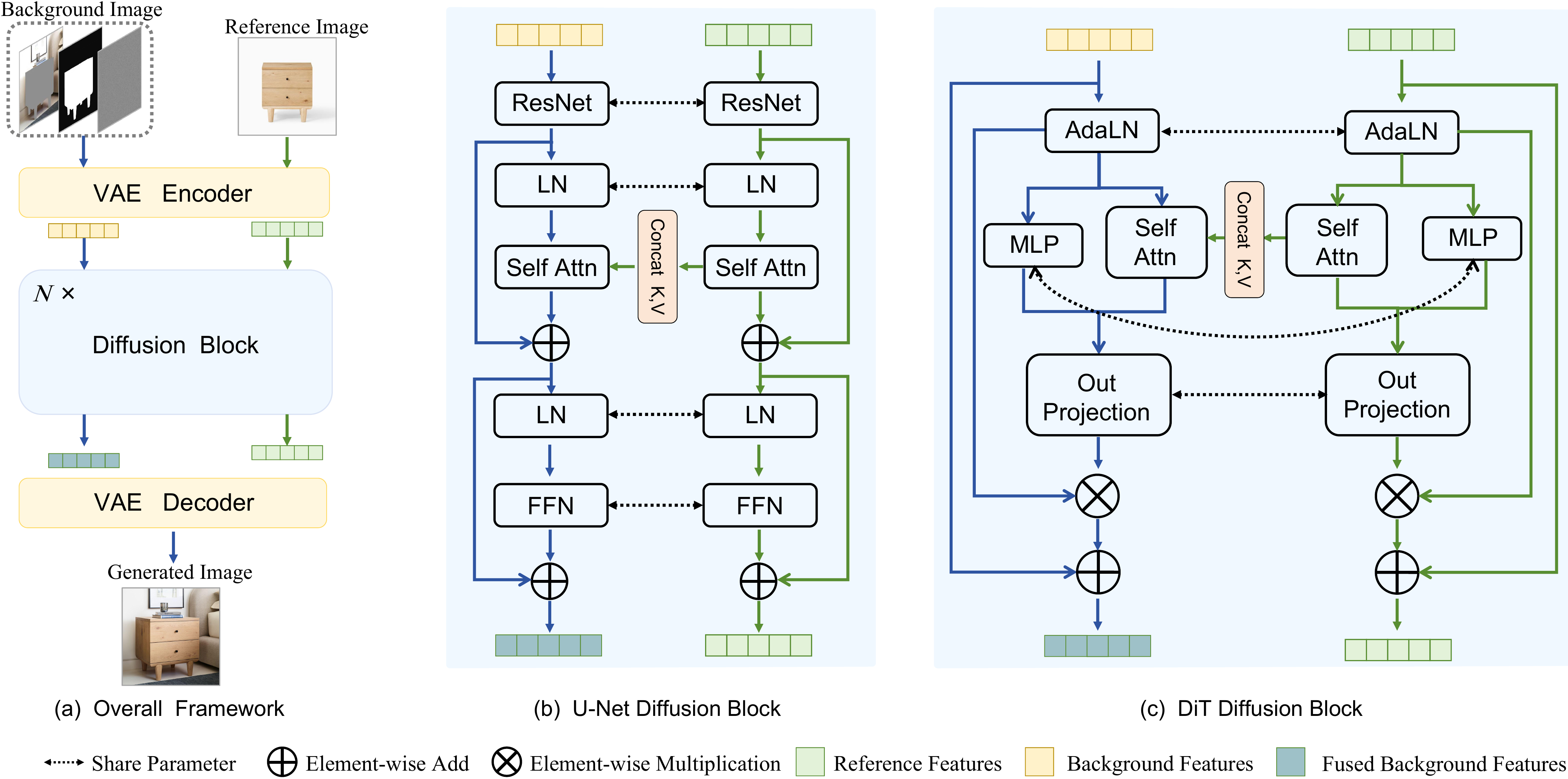} \caption{\textbf{The architecture of our RoomEditor++.} Our method shares parameters between the two diffusion backbones for unified feature space learning. As shown, reference features propagate independently, while background features interact with reference features through a self-attention module at each layer, ensuring effective feature alignment.} \label{fig:method} \end{figure*}

\section{RoomEditor++ for Furniture Synthesis}
\label{sec:method}
To address the issues of feature misalignment and structural incoherence lying in image composition, we present a RoomEditor++ architecture and show its underlying working mechanism. First, we elaborate on the core design of RoomEditor++, a parameter-sharing diffusion backbone compatible with both U-Net and DiT, which unifies reference feature extraction and background inpainting. Then, we provide a detailed mathematical analysis from the perspective of feature consistency, while comparing RoomEditor++ with state-of-the-arts to clarify its superiority. Finally, empirical validations are provided to verify the effectiveness of our design.

\subsection{RoomEditor++ Architecture}
The core of RoomEditor++ lies in its parameter-sharing diffusion backbone, which eliminates the feature domain gap between reference furniture and background scenes inherent in dual-branch methods. We first show the overall architecture design and then present its instantiations of U-Net and DiT backbones.

\subsubsection{\textbf{Overall Architecture}}
Given a masked background image $\bm{I}_{\mathrm{bg}}^{M}$ (with a predefined region for furniture placement) and a reference image $\bm{I}_{\mathrm{ref}}^{M}$ (containing the target furniture), the goal of RoomEditor++ is to seamlessly integrate the reference furniture into the masked region of the background while ensuring geometric alignment and texture coherence.
Unlike existing dual-branch diffusion methods (e.g., MimicBrush \cite{mimicbrush}) that employ two independent U-Nets for reference processing and background inpainting, RoomEditor++ adopts a shared diffusion backbone (instantiable as U-Net or DiT) to jointly process $\bm{I}_{\mathrm{ref}}^{M}$ and $\bm{I}_{\mathrm{bg}}^{M}$ within a unified parameter space. This design inherently enforces consistent feature representations across the two image domains.

As illustrated in \autoref{fig:method}, we build RoomEditor++ upon the inpainting diffusion model $\epsilon_{\theta}(\cdot)$ as the base network \cite{mimicbrush}, which has $N$ diffusion blocks. The shared backbone processes both $\bm{I}_{\mathrm{ref}}^{M}$ and $\bm{I}_{\mathrm{bg}}^{M}$ via the same parameter set. For the $l$-th interaction block (corresponding to a self-attention layer in either DiT or U-Net), the feature of $\bm{I}_{\mathrm{ref}}^{M}$, denoted as $f_{l}(\bm{I}_{\mathrm{ref}}^{M})$, is computed through self-attention:
\begin{equation}
      \label{ref}
      f_{l}(\bm{I}_\text{ref}^M) =  \mathrm{softmax} \left(\frac{\bm{Q}_\text{ref} \cdot \bm{K}_\text{ref}^{\top}}{\sqrt{d_k}}\right) \cdot \bm{V}_\text{ref},
      \end{equation}
where $\bm{Q}_\text{ref} = \mathcal{F}(f_{l-1}(\bm{I}_\text{ref}^M))\bm{W}_{Q}$, $\bm{K}_\text{ref} = \mathcal{F}(f_{l-1}(\bm{I}_\text{ref}^M))\bm{W}_{K}$, $\bm{V}_\text{ref} = \mathcal{F}(f_{l-1}(\bm{I}_\text{ref}^M))\bm{W}_{V}$, and $\bm{W}_{\{Q,K,V\}}$ are learnable projection matrices in the attention layer. $\mathcal{F}(\cdot)$ denotes post-attention operations, and $f_{l-1}(\bm{I}_\text{ref}^M)$ is the output of the $(l-1)$-th attention layer. Note that RoomEditor++ takes image, mask, and noise as inputs, and we omit mask and noise in the formulation for brevity.

A critical design of RoomEditor++ is the unidirectional feature interaction: reference features $f_{l-1}(\bm{I}_\text{ref}^M)$ are injected into background features $f_{l-1}(\bm{I}_\text{bg}^M)$ via a mixture attention mechanism to guide inpainting. Let $f_l(\cdot)$ denote the feature at the $l$-th attention layer of the shared backbone $\mathcal{B}$ (spatial feature map for U-Net, token sequence for DiT). The background feature $f_l(\bm{I}_\text{bg}^M)$ is computed as:
\begin{align}\label{equ:cross-att}
          f_{l}(\bm{I}_\text{bg}^M) &= \mathrm{softmax} \left(\frac{{\bm{Q}_{\text{bg}}} \cdot {\bm{K}_{[\text{bg},\text{ref}]}}^{\top}}{\sqrt{d_k}}\right) \cdot \bm{V}_{[\text{bg},\text{ref}]},
      \end{align}
where $\bm{Q}_{\text{bg}} =\mathcal{F} (f_{l-1}(\bm{I}_\text{bg}^M))\bm{W}_{Q}$, $\bm{K}_{[\text{bg},\text{ref}]} = \mathrm{cat}(\mathcal{F}(f_{l-1}(\bm{I}_\text{bg}^M)), \mathcal{F}(f_{l-1}(\bm{I}_\text{ref}^M)))\bm{W}_{K}$, $\bm{V}_{[\text{bg},\text{ref}]} = \mathrm{cat}(\mathcal{F}(f_{l-1}(\bm{I}_\text{bg}^M)), \mathcal{F}(f_{l-1}(\bm{I}_\text{ref}^M)))\bm{W}_{V}$, and $\mathrm{cat}(\cdot)$ denotes concatenation along the token dimension of DiT or spatial dimension of U-Net.
Notably, existing dual-backbone methods (e.g., MimicBrush \cite{mimicbrush}) rely on extra image encoders (e.g., CLIP \cite{clip}) for reference feature extraction. RoomEditor++ replaces such redundant modules with the mixture attention in Eqn.~(\ref{equ:cross-att}), resulting in a more concise and efficient architecture, regardless of whether $\mathcal{B}$ is U-Net or DiT.

For training {RoomEditor++}, we first construct the masked background as \(\bm{I}_\text{bg}^M = \bm{M}_\text{bg} \odot \bm{I}_{gt}\) using the ground\mbox{-}truth image \( \bm{I}_{\text{gt}} \) and a mask \( \bm{M}_\text{bg} \). We then sample random noise \( \epsilon \) and a timestep \( t \) to obtain the noisy image \( \bm{I}_{gt}^{t} \). Finally, we minimize the following loss:
\begin{equation}
\mathcal{L} = \mathbb{E}_{t, \bm{I}_{\text{gt}}, \epsilon} \left\| \epsilon_\theta(\bm{I}_{\text{gt}}^{t}, \bm{I}_\text{bg}^M, \bm{c}, t) - \epsilon \right\|_2^2,
\end{equation}
where \( \bm{c} \) denotes features extracted from \( \bm{I}_\text{ref}^M \) by the same backbone $\mathcal{B}$. This formulation is compatible with both U\mbox{-}Net and DiT denoisers.
For the instantiations of RoomEditor++ with U-Net and DiT, their distinction lies in $\mathcal{F}(\cdot)$ of the generic interaction operator $f_l(\cdot)$ in
Eqns.~(\ref{ref})--(\ref{equ:cross-att}).

\subsubsection{\textbf{U-Net Backbone Instantiation}}
\label{subsec:roomeditor_unet}
Generally, the $\mathcal{F}(\cdot)$ in the U\!-Net backbone denotes the post-attention
transformation module, consisting of a feed-forward network, a local
ResNet block, two layer-normalization operations, and the associated
residual connection.
As shown in ~\autoref{fig:method}(b), 
each block in the reference and background branches take
$\bm{x}_{l-1}^{\text{ref}}$ and $\bm{x}_{l-1}^{\text{bg}}$ as their inputs, respectively.

\vspace{0.2em}
\noindent\textit{ResNet \& layer normalization.}
Each branch first goes through a local ResNet block $R_l(\cdot)$
followed by layer normalization:
\begin{equation}
\tilde{\bm{x}}_l^{\,b}
=
\mathrm{LN}\!\bigl(R_l(\bm{x}_{l-1}^{\,b})\bigr),
\qquad
b \in \{\mathrm{ref},\mathrm{bg}\}.
\end{equation}

\vspace{0.2em}
\noindent\textit{Attention with concatenated keys and values.}
We then plug the normalized features $\tilde{\bm{x}}_l^\text{ref}$ and
$\tilde{\bm{x}}_l^\text{bg}$ into the interaction operator of
Eqn.~(\ref{ref}) and Eqn.~(\ref{equ:cross-att}). For the reference stream, Eqn.~(\ref{ref})
is evaluated with
\begin{equation*}
\bm{Q}_\text{ref} = \tilde{\bm{x}}_l^\text{ref} \bm{W}_Q,\quad
\bm{K}_\text{ref} = \tilde{\bm{x}}_l^\text{ref} \bm{W}_K,\quad
\bm{V}_\text{ref} = \tilde{\bm{x}}_l^\text{ref} \bm{W}_V,
\end{equation*}
where $\bm{W}_Q,\bm{W}_K,\bm{W}_V$ are learnable projection matrices. For the background stream we follow the mixture-attention operator in
Eqn.~(\ref{equ:cross-att}), where
\(\bm{Q}_\text{bg}=\tilde{\bm{x}}_l^\text{bg}\bm{W}_Q\),
\(\bm{K}_{[\text{bg},\text{ref}]}=[\tilde{\bm{x}}_l^\text{bg};\tilde{\bm{x}}_l^\text{ref}]\bm{W}_K\),
and
\(\bm{V}_{[\text{bg},\text{ref}]}=[\tilde{\bm{x}}_l^\text{bg};\tilde{\bm{x}}_l^\text{ref}]\bm{W}_V\),
i.e., Eqn.~(\ref{equ:cross-att}) is computed by using concatenated background and reference features. Then, the attention outputs are added back to the inputs:
\begin{equation}
\hat{\bm{x}}_l^{\,b}
=
\bm{x}_{l-1}^{\,b}
+
f_l(\bm{I}_{b}^{M}).
\qquad
\end{equation}
\vspace{0.2em}
\noindent\textit{FeedForward network.}
Finally, $\hat{\bm{x}}_l^{\text{ref}}$ and $\hat{\bm{x}}_l^{\text{bg}}$ are processed by a feed-forward network with a residual connection:

\begin{equation}
\bm{x}_l^{\,b} = \left( \max\left( 0, \mathrm{LN}(\hat{\bm{x}}_l^{\,b}) \bm{W}_1 \right) \bm{W}_2 \right) + \hat{\bm{x}}_l^{\,b},
\end{equation}
where \(\bm{W}_1\) and \(\bm{W}_2\) are the weight matrices for the two linear transformations in the feed-forward network. The resulting representations are then passed to the next U\!-Net block.

\subsubsection{\textbf{DiT Backbone Instantiation}}
\label{subsec:roomeditor_dit}
For DiT-based RoomEditor++, the interaction operators (Eqns.~(\ref{ref})–(\ref{equ:cross-att})) are applied to token sequences instead of spatial feature maps (\autoref{fig:method}(c)). Here, $\mathcal{F}(\cdot)$ includes the concatenation of attention and multilayer perceptron outputs, an output projection module, adaptive layer normalization (AdaLN)~\cite{Peebles2022DiT}, and residual connections. 
In each block, the reference and background branches take
$\bm{h}_{l-1}^\text{ref}$ and $\bm{h}_{l-1}^{\text{bg}}$ as their inputs, respectively.

\vspace{0.2em}
\noindent\textit{AdaLN-Zero.}
Each block first applies an AdaLN-Zero module conditioned on the diffusion
timestep embedding $\bm{\tau}$:
\begin{equation}
\tilde{\bm{h}}_l^{b},\; \bm{g}_l^{b}
=
\mathrm{AdaLNZero}\big(\bm{h}_{l-1}^{b},\, \bm{\tau}\big),
\qquad
b \in \{\text{ref},\text{bg}\},
\end{equation}
where $\tilde{\bm{h}}_l^{b}$ is the normalized sequence and
$\bm{g}_l^{b}\in\mathbb{R}^d$ is a learned gating vector.

\vspace{0.2em}
\noindent\textit{Attention with concatenated keys and values.}
By using $\tilde{\bm{h}}_l^{b}$ as input, the attention module strictly follows
the same interaction operators as Eqn.~(\ref{ref}) and Eqn.~(\ref{equ:cross-att}). For the
reference stream (self-attention), Eqn.~(\ref{ref}) is evaluated with
$\bm{Q}_\text{ref} = \tilde{\bm{h}}_l^\text{ref}\bm{W}_Q$,
$\bm{K}_\text{ref} = \tilde{\bm{h}}_l^\text{ref}\bm{W}_K$,
$\bm{V}_\text{ref} = \tilde{\bm{h}}_l^\text{ref}\bm{W}_V$.
For the background stream (mixture attention), Eqn.~(\ref{equ:cross-att}) is evaluated with
$\bm{Q}_\text{bg} = \tilde{\bm{h}}_l^\text{bg}\bm{W}_Q$,
$\bm{K}_{[\text{bg},\text{ref}]} =
[\tilde{\bm{h}}_l^\text{bg};\tilde{\bm{h}}_l^\text{ref}]\bm{W}_K$,
and
$\bm{V}_{[\text{bg},\text{ref}]} =
[\tilde{\bm{h}}_l^\text{bg};\tilde{\bm{h}}_l^\text{ref}]\bm{W}_V$,
where $\bm{W}_{\{Q,K,V\}}$ are learnable projection matrices.

\vspace{0.2em}
\noindent\textit{Single-stream Flux-style block.}
In the blocks that integrate image conditions, we further follow the
Flux single-stream design \cite{dehghani2023scaling} by running an MLP in parallel with the
attention branch and fusing them through a shared output projection.
For $b \in \{\text{ref},\text{bg}\}$, the MLP branch is
\begin{equation}
\bm{m}_l^{b} =
\phi\!\big(\tilde{\bm{h}}_l^{b} \bm{W}_m\big),
\end{equation}
where \(W_m\) is the learnable projection matrix of the MLP branch and $\phi(\cdot)$ denotes GELU. And the attention branch produces
$f_l(\bm{I}_b^M)$, which is instantiated as Eqn.~(\ref{ref}) for
$b=\text{ref}$ (self-attention) and as Eqn.~(\ref{equ:cross-att}) for
$b=\text{bg}$ (mixture attention).  The two branches are concatenated and
projected back to the model dimension:
\begin{equation}
\bm{u}_l^{b} =
\big[\,f_{l}(\bm{I}_{b}^M),\; \bm{m}_l^{b}\,\big] \bm{W}_o,
\end{equation}
where \(W_o\) is the shared output projection matrix that maps the
concatenated attention and MLP features back to the model dimension. The final output of the block is obtained via a gated residual
connection:
\begin{equation}
\bm{h}_l^{b}
=
\bm{h}_{l-1}^{b}
+
\bm{g}_l^{b} \odot \bm{u}_l^{b},
\end{equation}
where $\odot$ denotes a element-wise multiplication between the gating vector
$\bm{g}_l^{b}$ and the fused update $\bm{u}_l^{b}$; the resulting hidden
states $\bm{h}_l^{b}$ are then propagated to the next DiT layer. In this
way, the DiT-based RoomEditor++ still uses the same interaction operator
$f_l$ as Eqns.~(\ref{ref})--(\ref{equ:cross-att}), while benefiting from
the higher parallelism of the Flux-style single-stream Transformer
blocks.

\subsection{Analysis on Merit of RoomEditor++}
In this subsection, we compare with previous works~\cite{pbe,anydoor,mimicbrush} from the perspective of feature consistency, and show the potential advantage of our architecture in the context of high\mbox{-}fidelity furniture synthesis task.

\subsubsection{\textbf{Task Description}}
\label{task_description}
Since high\mbox{-}fidelity furniture synthesis aims to prominently place reference furniture $\bm{I}_{\text{ref}}$ in the background image $\bm{I}_{\text{bg}}$, it can be formulated as an ideal copy\mbox{-}paste problem. Specifically, the target image $\bm{I}_\text{gt}$ can be represented as a composition of a masked background $\bm{M}_{\text{bg}}\odot \bm{I}_{\text{bg}}$ and a masked reference $\bm{M}_{\text{ref}}\odot \bm{I}_{\text{ref}}$ with pose transformation $\mathcal{R}$:
\begin{equation}\label{equ:gt}
    \bm{I}_\text{gt}=\bm{M}_{\text{bg}}\odot \bm{I}_{\text{bg}}+\mathcal{R}(\bm{M}_{\text{ref}}\odot \bm{I}_{\text{ref}}),
\end{equation}
where $\odot$ indicates element\mbox{-}wise multiplication. Therefore, the key issue is learning the precise transformation $\mathcal{R}$, given a pair of background and reference images $\{\bm{I}_{\text{bg}}, \bm{I}_{\text{ref}}\}$. As $\mathcal{R}$ depends on the spatial and visual relationship between $\bm{I}_{\text{bg}}$ and $\bm{I}_{\text{ref}}$, the final objective can be formulated as 
\begin{equation}\label{equ:objectives}
\begin{aligned}
\bm{I}_\text{gt}
&=\bm{M}_{\text{bg}}\odot \bm{I}_{\text{bg}}
+\mathcal{R}(\bm{M}_{\text{ref}}\odot \bm{I}_{\text{ref}}\mid 
\bm{M}_{\text{bg}}\odot \bm{I}_{\text{bg}})\\
&= \bm{M}_{\text{bg}}\odot \bm{I}_{\text{bg}}
+ \overline{\bm{M}}_{\text{bg}}\odot
\mathcal{R}(\bm{I}_{\text{ref}}^{M}\mid\bm{I}_{\text{bg}}^{M}),
\end{aligned}
\end{equation}
where $\bm{I}_{\text{bg}}^{\text{M}}$ and $\bm{I}_{\text{ref}}^{\text{M}}$ represent $\bm{M}_{\text{bg}} \odot \bm{I}_{\text{bg}}$ and $\bm{M}_{\text{ref}} \odot \bm{I}_{\text{ref}}$, respectively, and $\overline{\bm{M}_{\text{bg}}} = \bm{1} - \bm{M}_{\text{bg}}$ denotes the complement of the mask $\bm{M}_{\text{bg}}$.

To learn transformation $\mathcal{R}$, diffusion model $\epsilon_{\theta}(\cdot)$ is generally employed to achieve this implicitly by training on abundant pairs of background and reference images $\{\bm{I}_{\text{bg}}, \bm{I}_{\text{ref}}\}$ with masks $\{\bm{M}_{\text{bg}}, \bm{M}_{\text{ref}}\}$.
Based on the empirical observation that inpainting models implicitly preserve unmasked regions during training, we treat predictions in these areas as ground truth. Accordingly, we consider two complementary cases: (1) masking the object region in $\bm{I}_{\text{gt}}$ to obtain $\bm{I}_{\text{bg}}^{\text{M}}$, where the background prediction remains unchanged; and (2) masking the background region to retain only the object (denoted as $\mathcal{R}(\bm{I}_{\text{ref}}^{\text{M}} \mid \bm{I}_{\text{bg}}^{\text{M}})$), where the object prediction remains unchanged. Combining these two masked predictions yields an approximation of the target image:
\begin{equation}
\label{equ:diffusion_app}
\begin{aligned}
\epsilon_{\theta}(\bm{I}_\text{gt}) 
&= \epsilon_{\theta}\!\left(\bm{M}_{\text{bg}}\odot\bm{I}_{\text{bg}}+ \overline{\bm{M}_{\text{bg}}}\odot\mathcal{R}\!\left(\bm{I}_{\text{ref}}^{\text{M}}\mid\bm{I}_{\text{bg}}^{\text{M}}\right)\right)\\
&\approx \bm{M}_{\text{bg}}\odot\epsilon_{\theta}\!\left(\bm{I}_{\text{bg}}^{\text{M}}\right) + \overline{\bm{M}_{\text{bg}}}\odot\epsilon_{\theta}\!\left(\mathcal{R}\!\left(\bm{I}_{\text{ref}}^{\text{M}}\mid \bm{I}_{\text{bg}}^{\text{M}}\right)\right).
\end{aligned}
\end{equation}

Similarly, since the inpainting model’s output in non-masked regions remains essentially unchanged regardless of input conditions, its impact at the feature level is marginal. Thus, for the $l$-th interaction block, we obtain
\begin{equation}
\label{equ:feature_app}
\begin{aligned}
f_{l}(\bm{I}_\text{gt}) 
&= f_{l}\!\left(\bm{M}_{\text{bg}}\odot\bm{I}_{\text{bg}} +\overline{\bm{M}_{\text{bg}}}\odot \mathcal{R}\!\left(\bm{I}_{\text{ref}}^{\text{M}}\mid \bm{I}_{\text{bg}}^{\text{M}}\right)\right) \\
&\approx \bm{M}_{\text{bg}}\odot f_{l}\!\left(\bm{I}_{\text{bg}}^{\text{M}}\right) +\overline{\bm{M}_{\text{bg}}}\odot f_{l}\!\left(\mathcal{R}\!\left(\bm{I}_{\text{ref}}^{\text{M}}\mid \bm{I}_{\text{bg}}^{\text{M}}\right)\right).
\end{aligned}
\end{equation}

Furthermore, the diffusion model implicitly learns $\mathcal{R}$ by progressively transforming and fusing features at each layer, rather than by directly transforming the original image. Thus, Eqn.~(\ref{equ:feature_app}) can be rewritten as follows in practice:
\begin{equation}\label{equ:feature_app2}
f_{l}\!\left(\bm{I}_{\text{gt}}\right) \approx \bm{M}_{\text{bg}}\odot f_{l}\!\left(\bm{I}_{\text{bg}}^{\text{M}}\right) + \overline{\bm{M}_{\text{bg}}}\odot\mathcal{R}_{l}\!\left(f_{l}\!\left(\bm{I}_{\text{ref}}^{\text{M}}\right)\mid f_{l}\!\left(\bm{I}_{\text{bg}}^{\text{M}}\right)\right),
\end{equation}
where $\mathcal{R}_{l}$ denotes the transformation in feature space. Given that the first term is solely related to the background, the differences among various methods~\cite{pbe,anydoor,mimicbrush} are primarily concentrated on the second term. According to Eqn.~(\ref{equ:diffusion_app}) and Eqn.~(\ref{equ:feature_app}), better feature consistency between ground-truth \( f_{l}(\bm{I}_{\text{gt}})\) and practice one (i.e., the right part of Eqn.~(\ref{equ:feature_app})) will help to optimize the inpainting model in~Eqn.~(\ref{equ:diffusion_app}). Empirical verification of Eqns.~(\ref{equ:diffusion_app}) and (\ref{equ:feature_app}) is shown in \autoref{verification_exp_for_5_6}.

\subsubsection{\textbf{Comparison with Previous Works}}
From the feature\mbox{-}consistency view, we compare \textit{RoomEditor++} with prior approaches~\cite{pbe,anydoor,mimicbrush}. Consider the $l$-th interaction block and how each method approximates $f_{l}(\bm{I}_{\text{gt}})$. Encoder\mbox{-}based methods~\cite{pbe,anydoor} use a frozen image encoder \(E(\cdot)\) and a trainable projector \(L(\cdot)\) to obtain reference features, formulated as $E_{L}(\bm{I}_\text{bg}^\text{M}) = L(E(\bm{I}_\text{bg}^\text{M}))$. Then, cross-attention is applied to inject the reference features \(L(E(\bm{I}_\text{ref}^\text{M})) \in \mathbb{R}^{T'\times d}\) into the background ones:
\begin{equation}
\label{equ:image_encoder_attn}
    \begin{aligned}
        \widehat{f_{l}(\bm{I}_{\text{gt}})}=\text{Cross-Attn}\big(\mathcal{F}(f_{l-1}(\bm{I}_\text{bg}^\text{M})),\, E_{L}(\bm{I}_\text{ref}^\text{M})\big),
    \end{aligned}
\end{equation}
where $\text{Cross-Attn}(\cdot,\cdot)$ denotes cross\mbox{-}attention (token\mbox{-}wise for DiT, and patch/spatial tokens for U\mbox{-}Net with attention blocks).

Dual\mbox{-}backbone methods (e.g.,~\cite{mimicbrush}) employ a frozen \emph{reference} backbone \( g_{l-1}(\cdot) \) to extract attention features at ($l$-1)-th interaction block and inject its features into the \emph{inpainting} backbone via a mixture of self\mbox{-} and cross\mbox{-}attention, expressed as
\begin{equation}
\label{equ:dual_attn}
\begin{aligned}
\widehat{f_{l}(\bm{I}_{\text{gt}})}
&\approx 
\underbrace{\text{Self-Attn}\!\big(\mathcal{F}(f_{l-1}(\bm{I}_\text{bg}^M))\big)}_{f_{l}(\bm{I}_\text{bg}^M)} \\
&\quad + 
\underbrace{\text{Cross-Attn}\!\big(\mathcal{F}(f_{l-1}(\bm{I}_\text{bg}^M)),\, \mathcal{G}(g_{l-1}(\bm{I}_\text{ref}^M))\big)}_{\bm{A}_{[\text{bg,ref}]}\!\big(\mathcal{G}(g_{l-1}(\bm{I}_\text{ref}^M))\big)\cdot \bm{W}_{V}},
\end{aligned}
\end{equation}
where $\text{Self-Attn}(\cdot,\cdot)$ denotes self-attention, $\mathcal{G}(\cdot)$ denotes the post-attention operations in the frozen
reference backbone and $\bm{A}_{[\text{bg,ref}]}$ is the attention score matrix. Compared to Eqn.~(\ref{equ:feature_app2}), these methods approximate \(\mathcal{R}_{l}\big(f_{l}(\bm{I}_\text{ref}^\text{M}) \mid f_{l}(\bm{I}_\text{bg}^\text{M})\big)\) through \(\bm{A}_{[\text{bg,ref}]}\big(\mathcal{G}(g_{l-1}(\bm{I}_\text{ref}^\text{M}))\big)\cdot \bm{W}_{V}\), while $\bm{M}_{\text{bg}}$ and $\overline{\bm{M}_{\text{bg}}}$ are implicitly learned through the attention mechanism.

In contrast, \textit{RoomEditor++} uses a \emph{single shared} backbone \( f_{l-1}(\cdot) \) to extract reference features and integrates them via mixture attention:
\begin{equation}
\label{equ:our_attn}
\begin{aligned}
\widehat{f_{l}(\bm{I}_{\mathrm{gt}})}
&\approx
\underbrace{\text{Self-Attn}\!\big(\mathcal{F}(f_{l-1}(\bm{I}_{\mathrm{bg}}^{M})\big)}_{\mathclap{f_{l}(\bm{I}_{\mathrm{bg}}^{M})}} \\
&\quad+
\underbrace{\text{Cross-Attn}\!\big(\mathcal{F}(f_{l-1}(\bm{I}_{\mathrm{bg}}^{M})),\, \mathcal{F}(f_{l-1}(\bm{I}_{\mathrm{ref}}^{M}))\big)}_{\mathclap{\bm{A}_{[\mathrm{bg,ref}]}\!\big(\mathcal{F}(f_{l-1}(\bm{I}_{\mathrm{ref}}^{M}))\big)\cdot \bm{W}_{V}}}.
\end{aligned}
\end{equation}

Compared to Eqn.~(\ref{equ:feature_app2}), our method approximates \(\mathcal{R}_{l}(f_{l}(\bm{I}_\text{ref}^\text{M}) \mid f_{l}(\bm{I}_\text{bg}^\text{M}))\) using \(\bm{A}_{[\text{bg,ref}]}(\mathcal{F}(f_{l-1}(\bm{I}_\text{ref}^\text{M}))) \cdot \bm{W}_{V}\). Under this formulation, we assume \(\mathcal{R}_{l}\) corresponds to a cross-attention operation, which applies a complex transformation to the reference features based on their similarity to the background features. It is worth note that the optimal \( g_{l}(\cdot) \) may be not \( f_{l}(\cdot) \), but \( f_{l}(\cdot) \) is a natural choice for offering the promising performance. Empirical comparisons in terms of loss visualizations and image quality indicate that shared\mbox{-}backbone \textit{RoomEditor++} outperforms both the frozen inpainting diffusion backbone and the unshared one. 

\begin{figure*}[t]
    \centering
    \includegraphics[width=0.9\textwidth]{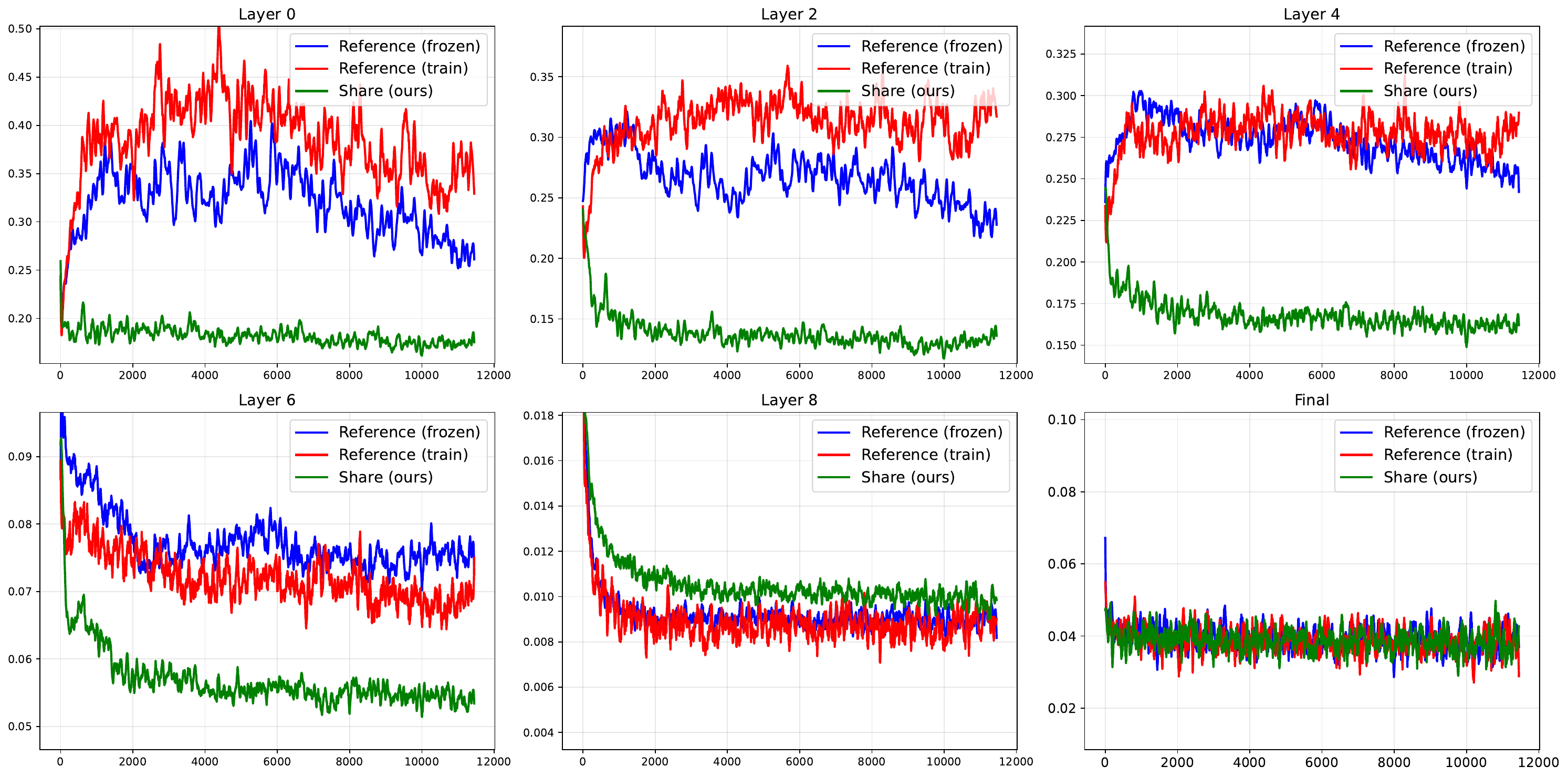}
    \caption{$\ell_2$ error across layers for three strategies: our \textbf{shared backbone} (U\mbox{-}Net; \textcolor{green}{green}), a \textbf{dual backbone} with \emph{frozen} reference (\textcolor{blue}{blue}), and a \textbf{dual backbone} with \emph{trainable} reference (\textcolor{red}{red}). ``Reference (train)'' means the reference backbone is frozen during training, as in \cite{mimicbrush}. ``Share (ours)'' indicates parameter sharing between inpainting and reference branches.}
    \label{fig:inter_compare_gt}
\end{figure*}

\subsection{Empirical Validation on Analysis}
To verify the above hypothesis, we further conducted experiments using a U-Net as the diffusion backbone.
\label{verification_exp_for_5_6}
\subsubsection{\textbf{Task Description}}
\label{empirical_task_description}

From \autoref{task_description}, we derived Eqn.~(\ref{equ:diffusion_app}) and Eqn.~(\ref{equ:feature_app}) based on the empirical properties of inpainting models and assumptions about the feature space. To more rigorously justify the validity of these assumptions and properties, we conducted the following experiments.

\begin{table}[!t]
\centering
\footnotesize  
\caption{Quantitative comparison between the average training loss and the region merging loss.}
\label{tab:exp_for_eqn_5}
\begin{tabular}{cc}
\toprule
\textbf{Training Loss (Avg.)} & \textbf{Region Merging Loss (Avg.)} \\
\midrule
$3.9\times10^{-2}$ & $4.3\times10^{-3}$ \\
\bottomrule
\end{tabular}
\end{table}

For Eqn.~(\ref{equ:diffusion_app}), the core assumption is that the inpainting diffusion model can preserve the unmasked regions of an image with high fidelity. Therefore, our goal is to demonstrate that the expression
\begin{equation}
\label{equ:validation_eq_5}
\begin{aligned}
\bm{M}_{\text{bg}} \odot \epsilon_{\theta}(\bm{I}_{\text{bg}}^{\text{M}}) 
+ \overline{\bm{M}_{\text{bg}}} \odot 
\epsilon_{\theta}(\mathcal{R}(\bm{I}_{\text{ref}}^{\text{M}} \mid \bm{I}_{\text{bg}}^{\text{M}})),
\end{aligned}
\end{equation}
can effectively approximate the ground truth (GT) denoising prediction $\epsilon_\theta(\boldsymbol{I_\mathrm{gt}})$.
To verify this assumption, we conduct experiments using a pretrained diffusion inpainting model on our test set. During the experiments, we add random noise to the input images and sample random timesteps $t$. The model is then provided with partial content from the full image (either the background or the object regions). We extract the predicted noise components from the corresponding regions and compute the $L_2$ loss as the evaluation metric:
\begin{equation}
\label{equ:validation_eq_6}
\begin{aligned}
\left\|\epsilon_\theta(\boldsymbol{I_\mathrm{gt}}) 
- \left(
\bm{M}_{\text{bg}} \odot \epsilon_{\theta}(\bm{I}_{\text{bg}}^{\text{M}}) 
+ \overline{\bm{M}_{\text{bg}}} \odot 
\epsilon_{\theta}(\mathcal{R}(\bm{I}_{\text{ref}}^{\text{M}} \mid \bm{I}_{\text{bg}}^{\text{M}}))
\right)
\right\|_2.
\end{aligned}
\end{equation}

As summarized in \autoref{tab:exp_for_eqn_5}, we can see that the merged result obtained by feeding the model with separated regions and combining their corresponding outputs yields a significantly lower loss than the average training loss (i.e., the average denoising loss during model training, visualized in \autoref{fig:inter_compare_gt}), thereby supporting the validity of the assumption in Eqn.~(\ref{equ:diffusion_app}).

\begin{table}[!t]
\centering
\footnotesize  
\setlength{\tabcolsep}{3pt}
\renewcommand{\arraystretch}{1.5}
\caption{\textbf{Cosine similarity across feature layers.} Here, \texttt{d}, \texttt{m}, and \texttt{u} correspond to the \textit{downsampling}, \textit{middle}, and \textit{upsampling} blocks of the network, respectively.}
\label{tab:feature_app}

\begin{tabular}{lcccccccc}
\toprule
\textbf{Layer}     & d0 & d1 & d2 & d3 & d4 & d5 & m0 & u0 \\
\midrule
\textbf{cos\_sim}  & 0.962 & 0.955 & 0.942 & 0.931 & 0.922 & 0.917 & 0.912 & 0.904 \\
\midrule
\textbf{Layer}     & u1 & u2 & u3 & u4 & u5 & u6 & u7 & u8 \\
\midrule
\textbf{cos\_sim} &  0.906 & 0.913 & 0.921 & 0.925 & 0.927 & 0.937 & 0.941 & 0.945 \\
\bottomrule
\end{tabular}

\end{table}

For Eqn.~(\ref{equ:feature_app}), the underlying assumption is that a property analogous to that in Eqn.~(\ref{equ:diffusion_app}) also holds at the feature level. To evaluate this hypothesis from a similarity perspective, we adopt \textit{cosine similarity} as the evaluation metric. Specifically, we compute the cosine similarity between the feature representations extracted from different attention layers on the validation set, defined as
\begin{equation}
\label{equ:validation}
\begin{aligned}
\text{cos}\Big(
f_{l}(\bm{I}_\text{gt}),\;
&\bm{M}_{\text{bg}}\odot f_{l}(\bm{I}_{\text{bg}}^{\text{M}})
\\
&+\overline{\bm{M}_{\text{bg}}}\odot 
f_{l}\!\big(\mathcal{R}(\bm{I}_{\text{ref}}^{\text{M}}\mid \bm{I}_{\text{bg}}^{\text{M}})\big)
\Big),
\end{aligned}
\end{equation}
where $f_l(\cdot)$ denotes the feature map extracted from the $l$-th layer of the model. The results in \autoref{tab:feature_app} show that cosine similarities remain consistently above 0.9 across most layers, supporting the validity of the assumption in Eqn.~(\ref{equ:feature_app}) and indicating that the feature-level composition effectively preserves the representational coherence of the full image.

\subsubsection{\textbf{Comparison with Previous Works}}
To further investigate the effect of feature consistency, we compare three variants on our test set: (i) a dual branch with a \emph{frozen} reference backbone as in MimicBrush~\cite{mimicbrush}, (ii) a dual branch with a \emph{trainable} reference backbone, and (iii) our \textit{RoomEditor++} with a \emph{shared} backbone for background and reference. In this study we instantiate the denoiser backbone as a U\mbox{-}Net and train all models under exactly the same setting, then compute the $\ell_2$ loss between features from two separate inputs ($\widehat{f_{l}(\bm{I}_{\text{gt}})}$) and those of the ground-truth images, which are outputted by the intermediate layers of inpainting U-Net (${f_{l}(\bm{I}_{\text{gt}})}$). From \autoref{fig:inter_compare_gt}, we can conclude that: 
(1) Our shared\mbox{-}backbone {RoomEditor++} (\textcolor{green}{green}) maintains consistently low $\ell_2$ loss across all intermediate layers. Notably, it achieves substantially lower $\ell_2$ loss than the other two methods at the early stages.
(2) The frozen (\textcolor{blue}{blue}) and trainable (\textcolor{red}{red}) reference branches approximate the features of ground\mbox{-}truth images only at the final stage, where they are strongly constrained by the training loss.
(3) The frozen reference branch outperforms the trainable one in the early stages, indicating that the \emph{unshared} dual\mbox{-}branch is harder to optimize due to large parameters and limited data.

\begin{remark} According to Eqn.~(\ref{equ:diffusion_app}) and Eqn.~(\ref{equ:feature_app}) in conjunction with the results in~\autoref{fig:inter_compare_gt}, we can observe the \emph{parameter-sharing dual backbone}, rather than a U\mbox{-}Net\mbox{-}specific inductive bias, is the main driver of improved feature consistency and downstream performance (see~\autoref{tab:ablation_study_indoor_bench}).  
\end{remark}

Although implemented with a U\mbox{-}Net, the formulation is backbone\mbox{-}agnostic and extends directly to DiT, where spatial features are replaced by token sequences while preserving the same objective, interaction, and shared\mbox{-}backbone advantages.

\begin{table*}[t]
\centering
\caption{\textbf{Quantitative comparison on realistic-scene subset of RoomBench++.} AnyDoor, MimicBrush and DreamFuse are evaluated under two settings: (\emph{i}) released models on large datasets and (\emph{ii}) fine-tuned on RoomBench++ (marked with $\star$). 
}
\label{tab:main_roombenchv1}

\begin{tabularx}{0.9\textwidth}{l|c|c|*{6}{>{\centering\arraybackslash}X}}
\toprule
Method & \makecell{Train\\Data$^{\ddagger}$} & RoomBench++ & FID↓ & SSIM↑ & PSNR↑ & LPIPS↓ & CLIP↑ & DINO↑ \\
\midrule
AnyDoor~\cite{anydoor}   & 500K  & \ding{55} & 28.03 & 0.767 & 18.85 & 0.137 & 87.74 & 76.91 \\
MimicBrush~\cite{mimicbrush} & 400K & \ding{55} & 22.50 & 0.784 & 19.41 & 0.111 & 88.21 & 79.11 \\
DreamFuse~\cite{huang2025dreamfuse} & 80k & \ding{55} & 18.50 & 0.793 & 21.45 & 0.099 & 91.34 & 89.16 \\
\midrule
AnyDoor$^\star$          & 500K  & \ding{51}(Full set) & 24.71 & 0.787 & 19.36 & 0.134 & 89.12 & 81.67 \\
MimicBrush$^\star$       & 400K  & \ding{51}(Full set) & 17.58 & 0.826 & 20.44 & 0.098 & 90.87 & 85.65 \\
DreamFuse$^\star$              & 80k  & \ding{51}(Full set) & 16.85 & 0.835 & \underline{22.23} & 0.091 & \textbf{92.28} & \underline{90.13} \\
\midrule
RoomEditor \cite{lin2025roomeditor} & 0 & \ding{51}(Realistic subset) & 18.42 & 0.793 & 21.15 & 0.094 & 90.51 & 85.47 \\
\midrule
RoomEditor++ (U)  & 0  & \ding{51}(Full set) & \underline{16.22} & \underline{0.842} & 22.07 & \underline{0.089} & 91.38 & 87.53 \\
RoomEditor++    & 0  & \ding{51}(Full set) & \textbf{15.88} & \textbf{0.862} & \textbf{22.96} & \textbf{0.085} & \underline{91.79} & \textbf{90.42} \\

\bottomrule
\end{tabularx}

\end{table*}

\begin{table*}[t]
\caption{\textbf{Quantitative comparison on full RoomBench++.} AnyDoor, MimicBrush and DreamFuse are evaluated under two settings: 
(\emph{i}) released models on large datasets and (\emph{ii}) fine-tuned on RoomBench++ (marked with $\star$).
}
\label{tab:main_roombenchv2}
\centering
\begin{tabularx}{0.9\textwidth}{l|c|c|*{6}{>{\centering\arraybackslash}X}}
\toprule
Method & \makecell{Train\\Data$^{\ddagger}$} & RoomBench++ & FID↓ & SSIM↑ & PSNR↑ & LPIPS↓ & CLIP↑ & DINO↑ \\
\midrule
AnyDoor~\cite{anydoor}   & 500K  & \ding{55} & 22.84 & 0.837 & 22.08 & 0.107 & 84.87 & 75.40 \\
MimicBrush~\cite{mimicbrush} & 400K & \ding{55} & 23.34 & 0.828 & 21.49 & 0.117 & 83.87 & 74.62 \\
DreamFuse~\cite{huang2025dreamfuse} & 80k  & \ding{55} & 20.30 & 0.844 & 23.91 & 0.102 & 87.92 & 87.63 \\

\midrule
AnyDoor$^\star$          & 500K  & \ding{51}(Full set) & 19.55 & 0.858 & 22.91 & 0.103 & 86.49 & 78.84 \\
MimicBrush$^\star$       & 400K  & \ding{51}(Full set) & 14.54 & 0.870 & 24.46 & 0.087 & 88.37 & 85.37 \\
DreamFuse$^\star$        & 80k  & \ding{51}(Full set) & 12.67 & 0.886 & 25.52 & 0.079 & 90.67 & \underline{89.15} \\
\midrule
RoomEditor \cite{lin2025roomeditor} & 0 & \ding{51}(Realistic subset) & 14.78 & 0.859 & 24.75 & 0.078 & 89.43 & 83.73 \\
\midrule
RoomEditor++ (U) & 0  & \ding{51}(Full set) & \underline{11.93} & \underline{0.891} & \underline{26.10} & \underline{0.071} & \underline{90.99} & 87.17 \\
RoomEditor++    & 0  & \ding{51}(Full set) & \textbf{11.49} & \textbf{0.905} & \textbf{26.82} & \textbf{0.067} & \textbf{91.39} & \textbf{91.03} \\
\bottomrule
\end{tabularx}
\end{table*}

\section{Experiments}
In this section, we first describe experimental setup. Then, we compare our method with existing approaches (i.e., encoder-based AnyDoor~\cite{anydoor}, Dual U-Net-based MimicBrush~\cite{mimicbrush} and DiT-based DreamFuse~\cite{huang2025dreamfuse}) by conducting quantitative and qualitative evaluations on our RoomBench++, 3D-FUTURE~\cite{3d-future}, and DreamBooth~\cite{dreambooth} datasets. Finally, we perform ablation studies to validate the effectiveness of both dataset configuration and model architectures.

\label{sec:exp}
\subsection{Experimental Setup}
\label{sec:exp_setup}
\noindent\textbf{Implementation Details.}  
\begin{table}[!t]
    \caption{Image and Mask Augmentation Strategies.}
    \label{tab:augmentation}
    \centering
    \footnotesize
    \renewcommand{\arraystretch}{1.1}  

    \begin{tabular}{l l p{0.50\linewidth}} 
        \hline
        \textbf{Type} & \textbf{Aug.} & \textbf{Description} \\  
        \hline

        \multirow{4}{*}{Image} 
        & Horiz.\ Flip & Probability 50\%. \\  
        & Rotation & Probability 50\%, angle $\le 30^\circ$. \\  
        & Scaling & Probability 30\%, range $\pm 20\%$. \\  
        & Cropping & Minimum retained ratio: 0.75. \\  
        \hline

        \multirow{4}{*}{Mask}  
        & Perturb.\ & 25\%: Random dilation/erosion. \\  
        & Blurring & 25\%: Boundary smoothing. \\  
        & BBox & 25\%: Replace mask by bounding box. \\  
        & None & 25\%: Keep original mask. \\  
        \hline
    \end{tabular}
\end{table}
We provide two RoomEditor++ models, where RoomEditor++(U) is constructed by U-Net backbone ($N=16$) initialized with Stable Diffusion 1.5 inpainting model~\cite{stable_diffusion}, and RoomEditor++ is constructed by DiT backbone ($N=57$) initialized with the FLUX.1 Fill [dev] \cite{flux2024}. 
Both models are trained on the RoomBench++ dataset for 5k steps with a batch size of 32 across four NVIDIA A800 GPUs, fine-tuned using LoRA~\cite{hu2022lora} with a rank of 256. We use the Prodigy optimizer~\cite{mishchenko2023prodigy} with safeguard warmup and bias correction enabled, applying a weight decay of 0.01 and set the input resolution to \(768 \times 768\). To enhance robustness against imperfect mask inputs and domain variations, we apply augmentation techniques to both masks and images. The details of mask augmentation are shown in \autoref{tab:augmentation}. 

\noindent\textbf{Objective Evaluation Metrics.}
To compare with other methods, we use SSIM \cite{ssim} and PSNR~\cite{psnr} for assessing reconstruction quality, FID~\cite{fid} and LPIPS~\cite{lpips} for assessing perceptual realism, CLIP-score~\cite{clip} and DINO-score~\cite{dinov2} for assessing semantic consistency.

\noindent\textbf{Human Evaluation Metrics.}
To complement objective metrics, we conducted a user study following \cite{mimicbrush}, evaluating results in fidelity (retaining reference identity and details), harmony (seamless integration with the background), and quality (overall visual appeal and detail).

\begin{figure*}[t]
    \centering
    \includegraphics[width=0.96\textwidth]{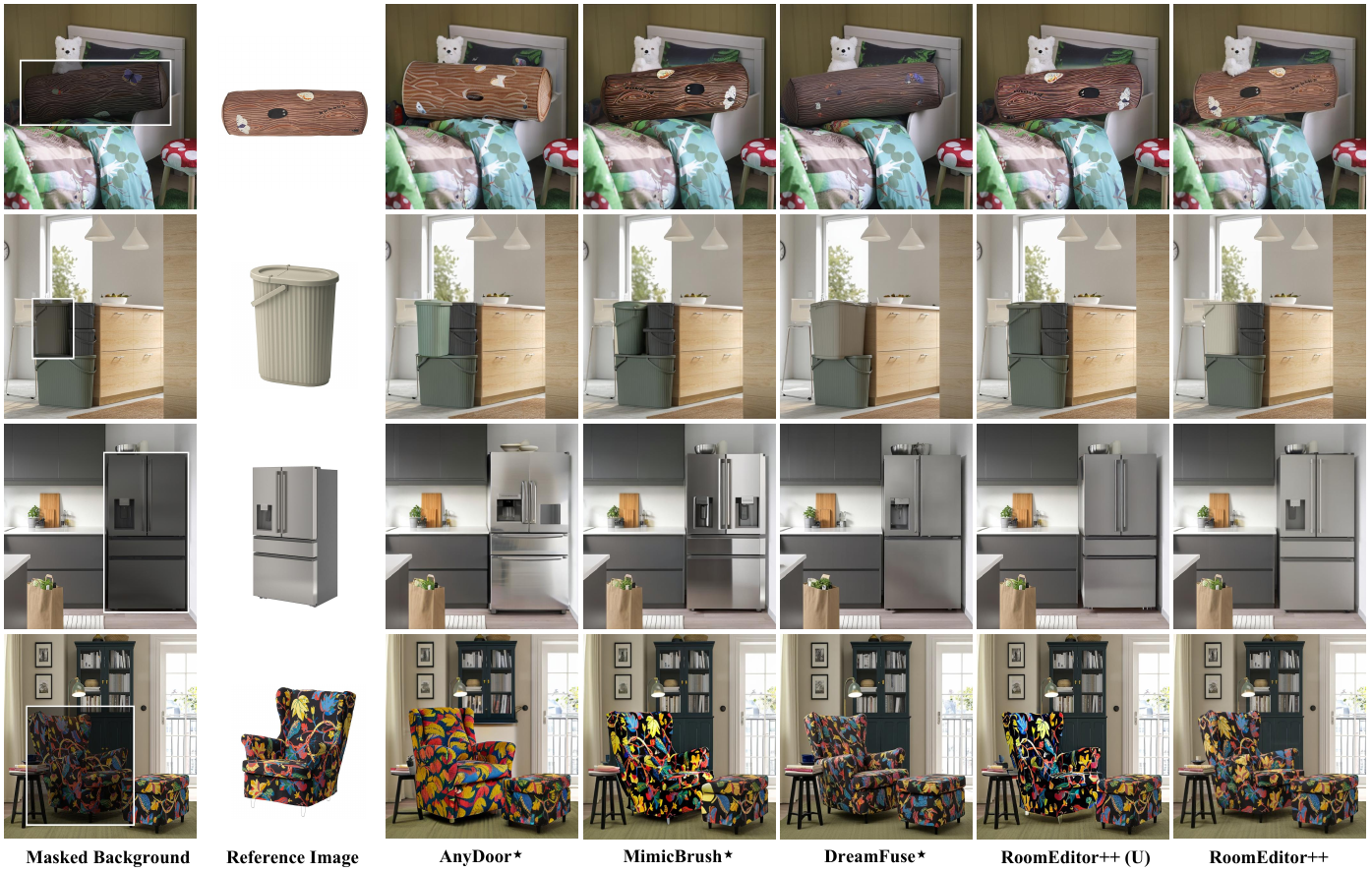}
    \caption{\textbf{Qualitative comparison on realistic-scene data.}
    RoomEditor++ consistently preserves identity and maintains visual coherence compared to existing methods after fine-tuning, including AnyDoor~\cite{anydoor}, MimicBrush~\cite{mimicbrush} and DreamFuse~\cite{huang2025dreamfuse}.} 
    \label{fig:indoor_bench_picture}
\end{figure*}

\begin{figure*}[!t]
    \centering
    \includegraphics[width=0.96\textwidth]{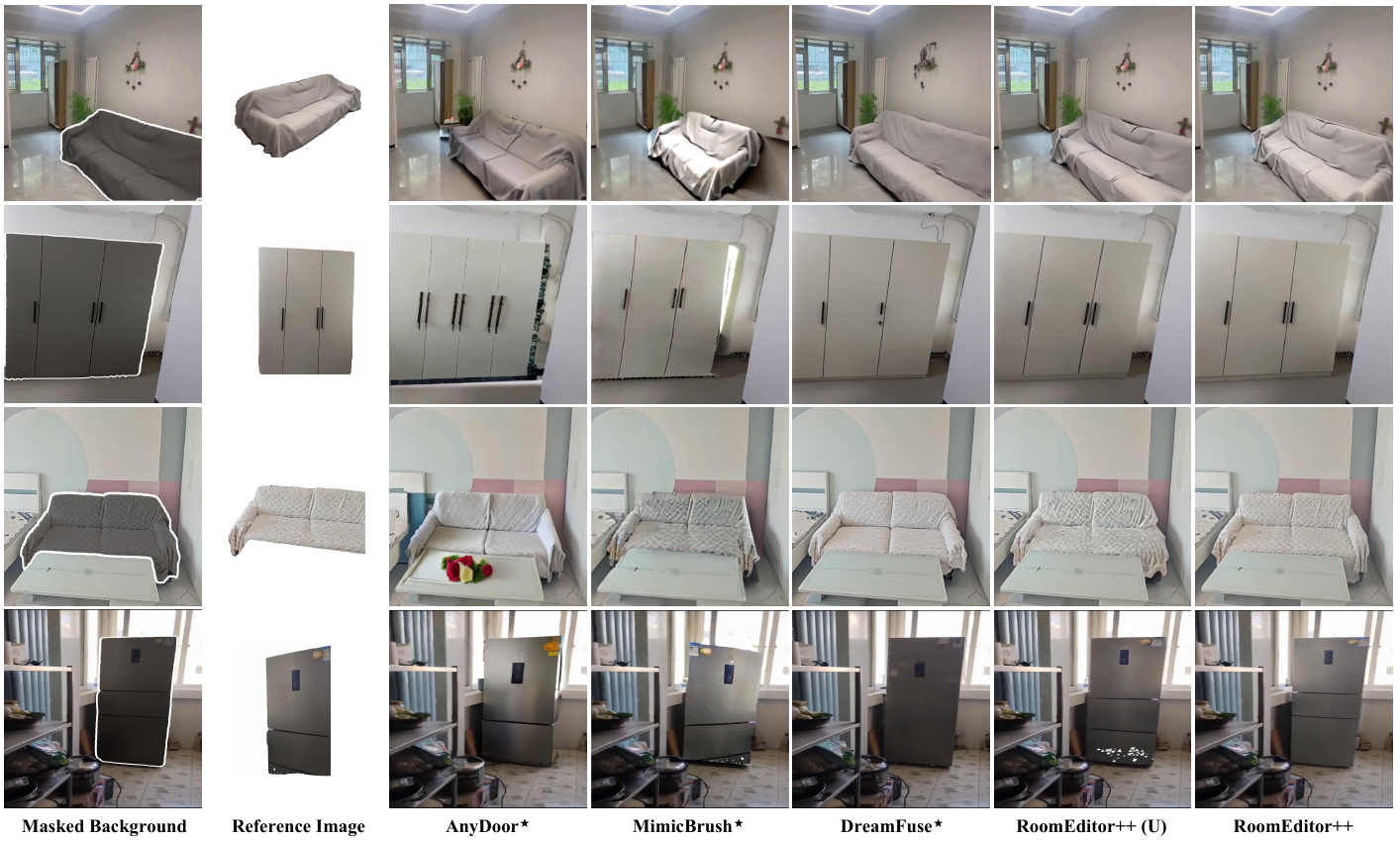}
    \caption{\textbf{Qualitative comparison on real-scene data.}
    RoomEditor++ better preserves the identity of the target object and produces more natural results compared with fine-tuned existing methods, including AnyDoor~\cite{anydoor}, MimicBrush~\cite{mimicbrush}, and DreamFuse~\cite{huang2025dreamfuse}.
 } 
    \label{fig:indoor_roombenchv2}
\end{figure*}
\subsection{Results on RoomBench++}
We first compare our RoomEditor++ with several competing methods (i.e., AnyDoor~\cite{anydoor}, Mimicbrush~\cite{mimicbrush} and DreamFuse~\cite{huang2025dreamfuse}) on our RoomBench++ under two settings, where the competing models are trained on their respective large datasets or further fine-tuned on our RoomBench++ (denoted with $\star$). Unless otherwise specified, RoomEditor++ refers to the DiT variant throughout the remainder of the experiments.

\subsubsection{\textbf{Quantitative Comparison}}
\autoref{tab:main_roombenchv1} and \autoref{tab:main_roombenchv2} respectively  give the results of different methods on  the 895 realistic-scene image pairs from RoomBench++ (i.e., the same RoomBench as in~\cite{lin2025roomeditor}) and the full testing set in RoomBench++ dataset, and the latter involves both real-scene images and realistic-scene images. From them we can draw the following conclusions: (1) The original models of AnyDoor, MimicBrush, and DreamFuse exhibit limited performance in furniture synthesis despite being trained on large-scale datasets. Notably, the large-scale datasets used for training AnyDoor and MimicBrush primarily consist of video data, highlighting the necessity of establishing a specific dataset tailored for home design. (2) Fine-tuning of the competing methods on our RoomBench++ significantly improves the performance of AnyDoor$^\star$, MimicBrush$^\star$, and DreamFuse$^\star$ across all metrics. In particular, the largest improvement is observed in the FID score on the RoomBench++ test set, with improvements of 3.29, 8.80, and 7.63, respectively. This also validates the effectiveness of RoomBench++. (3) RoomEditor++ (U) clearly outperforms our previous work (i.e., RoomEditor \cite{lin2025roomeditor}) by using the same model architecture, showing the effectiveness of our RoomBench++. (4) Our RoomEditor++ (U) and RoomEditor++, which are trained solely on RoomBench++, outperform competing methods in most quantitative metrics. Besides, performance of RoomEditor++ (U) lags behind that of the DiT-based RoomEditor++, indicating the effectiveness of DiT architecture. These results above clearly demonstrate that the significance of our RoomBench++, and our parameter-sharing diffusion backbone provides an effective solution for furniture synthesis, especially for unavailable large-scale training data.

\subsubsection{\textbf{User Study}}
Following the protocol in~\cite{mimicbrush}, we conduct a user study to evaluate the perceptual quality of generated images. We let 20 annotators (11 undergraduate students, 5 parents of some of the students, and 4 university faculty or staff members, all informed of the evaluation criteria) rank one hundred randomly selected generation results of different methods based on our benchmark from three aspects: fidelity, harmony, and overall quality. As shown in \autoref{tab:user_study}, our RoomEditor++ outperforms competing methods across all aspects, which further verify the effectiveness of our work.

\begin{table}[!t]
\caption{\textbf{User study results.} In each trial, annotators were presented with four images (one from each method) and asked to rank them with respect to fidelity, harmony, and overall quality. 
``Best (\%)'' denotes the percentage of cases where a method was ranked first (i.e., perceived as the best under the given criterion), while ``Rank↓'' indicates the average ranking (lower is better) across all trials.}
\centering

\begin{tabularx}{0.48\textwidth}{l *{6}{>{\centering\arraybackslash}X}}
\toprule
Method &
{\scriptsize \shortstack{Fidelity\\Best (\%)}} &
{\scriptsize \shortstack{Fidelity\\Rank↓}} &
{\scriptsize \shortstack{Harmony\\Best (\%)}} &
{\scriptsize \shortstack{Harmony\\Rank↓}} &
{\scriptsize \shortstack{Quality\\Best (\%)}} &
{\scriptsize \shortstack{Quality\\Rank↓}} \\
\midrule
AnyDoor$^{\star}$~\cite{anydoor}  & 8.1 & 3.42 & 9.4 & 3.38 & 8.0  & 3.43 \\
MimicBrush$^{\star}$~\cite{mimicbrush} & 8.3 & 3.39 & 7.6 & 3.44 & 9.9 & 3.40 \\
DreamFuse$^{\star}$~\cite{huang2025dreamfuse} & 22.1 & 2.55 & 27.4 & 2.22 & 23.9 & 2.52 \\
\midrule
RoomEditor++ (U) & 26.0 & 2.27 & 24.6 & 2.34 & 25.0 & 2.40 \\
RoomEditor++  & \textbf{35.5} & \textbf{1.69} & \textbf{31.0} & \textbf{1.89} & \textbf{33.2} & \textbf{1.77} \\
\bottomrule
\end{tabularx}
\label{tab:user_study}
\end{table}

\subsubsection{\textbf{Qualitative Comparison}}
As illustrated in \autoref{fig:indoor_bench_picture} (Rows 2 and 4) and \autoref{fig:indoor_roombenchv2} (Row 2), AnyDoor$^\star$~\cite{pbe} fails to keep object identity, leading to significant deviation from the reference. \autoref{fig:indoor_bench_picture} (Rows 2 and 4) and \autoref{fig:indoor_roombenchv2} (Row 1) indicate that MimicBrush$^\star$~\cite{anydoor} better preserves identity but faces issues of improper scaling, color mismatches, and misplacement. Although DreamFuse$^\star$~\cite{huang2025dreamfuse} achieves relatively high harmony, but its fidelity is insufficient, as shown in \autoref{fig:indoor_bench_picture} (Rows 2 and 3) and \autoref{fig:indoor_roombenchv2} (Row 4). RoomEditor++ (U) strikes a balance between fidelity and harmony, but remains suboptimal in both aspects. In contrast, RoomEditor++ achieves seamless integration, realism, and correct orientation. Consequently, our RoomEditor++ shows better generation quality in furniture synthesis.
\begin{table}[!t]
\caption{\textbf{Cross-dataset evaluation on 3D-FUTURE~\cite{3d-future}}. Our model outperforms MimicBrush across multiple metrics, demonstrating strong generalization to unseen data distributions.}
\centering
\begin{tabularx}{0.48\textwidth}{l *{6}{>{\centering\arraybackslash}X}}
\toprule
Method &
{\scriptsize FID↓} &
{\scriptsize SSIM↑} &
{\scriptsize PSNR↑} &
{\scriptsize LPIPS↓} &
{\scriptsize CLIP↑} &
{\scriptsize DINO↑} \\
\midrule
Anydoor~\cite{anydoor} & \underline{12.34} & \underline{0.797} & \underline{22.65} & \underline{0.131} & 82.64 & 71.71 \\
MimicBrush~\cite{mimicbrush} & 14.20 & 0.660 & 20.64 & 0.272 & 80.10 & 64.61 \\
DreamFuse~\cite{huang2025dreamfuse} & 14.59 & 0.667 & 22.22 & 0.252 & \underline{87.60} & \underline{83.40} \\
RoomEditor~\cite{lin2025roomeditor} & 13.91 & 0.658 & 20.93 & 0.260 & 80.28 & 66.97 \\ 
\midrule
RoomEditor++ & \textbf{5.03} & \textbf{0.826} & \textbf{26.48} & \textbf{0.110} & \textbf{88.92} & \textbf{88.07} \\
\bottomrule
\end{tabularx}

\label{tab:3dfuture}
\end{table}

\begin{figure*}[!t]
    \centering
    \includegraphics[width=0.96\textwidth]{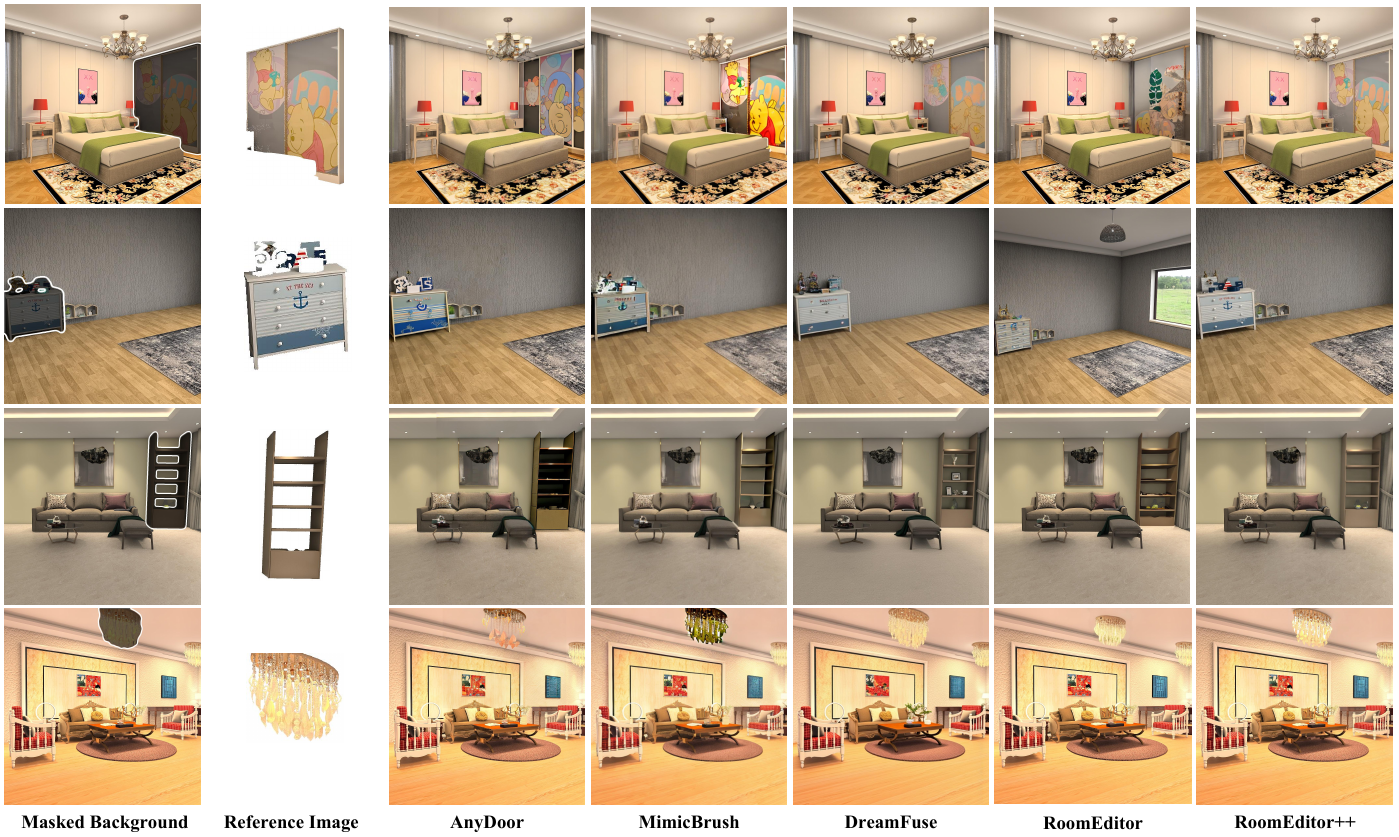}
    \caption{\textbf{Qualitative comparison on 3D-Future data.} RoomEditor++ performs exceptionally well in capturing the patterns of reference images with high fidelity (rows 1 and 2). It also maintains excellent consistency and harmony with the edited images (rows 3 and 4) compared with existing methods, including AnyDoor~\cite{anydoor}, MimicBrush~\cite{mimicbrush}, DreamFuse~\cite{huang2025dreamfuse} and RoomEditor~\cite{lin2025roomeditor}.
}
    \label{fig:compare_3dfuture}
\end{figure*}

\subsection{Generalization to 3D-FUTURE}
To evaluate the generalization ability of our method on unseen datasets, we conducted cross-dataset
experiments using the 3D-FUTURE~\cite{3d-future} dataset, which differs from our RoomBench++ in visual domain and object composition.
We selected 1,020 samples covering 34 furniture categories (randomly 30 samples per category).
As the dataset does not provide paired reference images, we generated pseudo-pairs by horizontally flipping reference objects and applying Gaussian-blurred masks to prevent trivial copy-paste solutions.
As shown in \autoref{tab:3dfuture}, despite being trained exclusively on RoomBench++, our model achieves superior performance in terms of all the metrics. In particular, RoomEditor++ achieves improvements of 7.31 and 4.67 points in FID and DINO metrics, respectively, compared to the suboptimal models AnyDoor and DreamFuse. Qualitative results in \autoref{fig:compare_3dfuture} further demonstrate strong generalization to unseen furniture data and background scenes without the need for fine-tuning.

\begin{table}[!t]
\caption{\textbf{Quantitative evaluation on the DreamBooth dataset~\cite{dreambooth}}. We compare our method with existing approaches. Despite being trained only on RoomBench, our method achieves the best results, showcasing strong generalization capability.}
\label{tab:main_exp_dreambooth}
\centering

\begin{tabularx}{0.48\textwidth}{l *{6}{>{\centering\arraybackslash}X}}
\toprule
Method &
{\scriptsize FID↓} &
{\scriptsize SSIM↑} &
{\scriptsize PSNR↑} &
{\scriptsize LPIPS↓} &
{\scriptsize CLIP↑} &
{\scriptsize DINO↑} \\
\midrule
AnyDoor~\cite{anydoor}       & 86.07  & 0.567 & 14.51 & 0.372 & 86.41 & 82.67 \\
MimicBrush~\cite{mimicbrush} & 77.16  & 0.585 & 15.35 & 0.361 & 87.66 & 83.70 \\
DreamFuse~\cite{huang2025dreamfuse}  & \underline{58.98} & 0.591 & \textbf{19.34} & \underline{0.278} & \underline{92.42} & \textbf{88.10} \\
RoomEditor~\cite{lin2025roomeditor} & 68.78 & \underline{0.594} & 16.43 & 0.304 & 89.62 & 85.04 \\
\midrule
RoomEditor++          & \textbf{57.64} & \textbf{0.728} & \underline{18.57} & \textbf{0.201} & \textbf{92.51} & \underline{87.76} \\
\bottomrule
\end{tabularx}

\end{table}
\begin{figure*}[!t]
    \centering
    \includegraphics[width=0.96\textwidth]{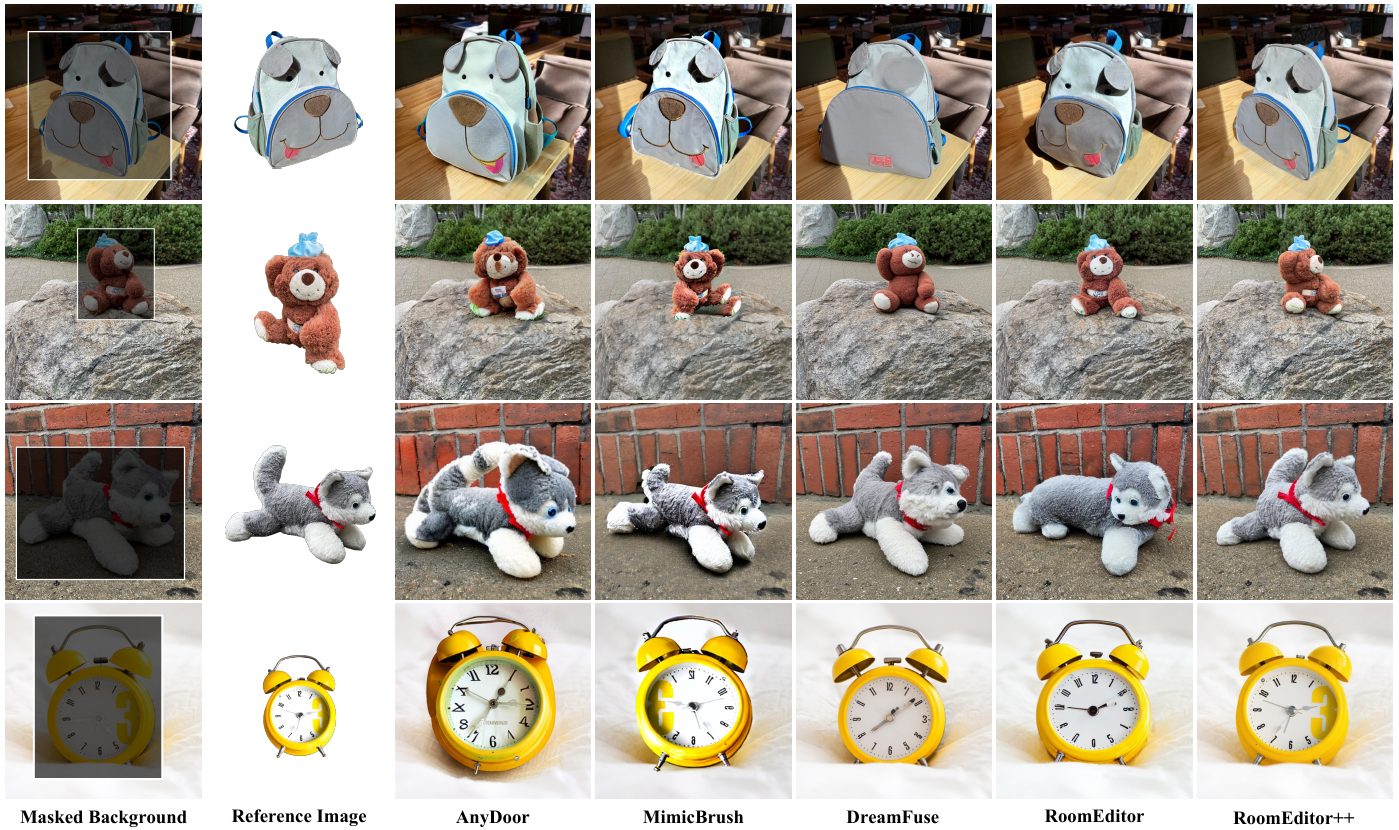}
    \caption{\textbf{Qualitative comparison on DreamBooth data}. RoomEditor++ demonstrates superior contextual coherence (rows 1 and 2) and higher fidelity (rows 3 and 4) compared with existing methods, including AnyDoor~\cite{anydoor}, MimicBrush~\cite{mimicbrush}, DreamFuse~\cite{huang2025dreamfuse} and RoomEditor~\cite{lin2025roomeditor}.}
    \label{fig:indoor_dreambooth}
\end{figure*}
\subsection{Generalization to DreamBooth}
We evaluate the generalization performance of our model using the DreamBooth dataset~\cite{dreambooth}, which contains 30 object categories, including backpacks, animals, sunglasses, and cartoon characters. Each sample is annotated with object masks, consistent with the realistic-scene data in the RoomBench++ dataset. In this experiment, the RoomEditor++ model is neither re-trained nor fine-tuned on external datasets. We compare it with publicly available models from~\cite{pbe, anydoor, mimicbrush}. As shown in \autoref{tab:main_exp_dreambooth}, our method significantly outperforms existing approaches trained on large-scale datasets on most metrics. In particular, DreamFuse achieves the highest PSNR-score and DINO-score, which indicates it has a certain advantage over other methods in terms of semantic understanding. These results demonstrate the strong generalization ability of RoomEditor++ in synthesizing realistic object placements in unseen environments, indicating its robustness for real-world applications. As shown in ~\autoref{fig:indoor_dreambooth}, AnyDoor~\cite{anydoor} and MimicBrush~\cite{mimicbrush} achieve reasonable fidelity, but their harmony is suboptimal and exhibits noticeable deviations from the surrounding context (e.g., Rows 1 and 3). DreamFuse~\cite{huang2025dreamfuse} demonstrates strong harmony but fails to faithfully align with the reference image (e.g., Rows 1 and 2). RoomEditor~\cite{lin2025roomeditor} performs reasonably well in both fidelity and harmony, yet still lags behind RoomEditor++.

\subsection{Ablation Study} 
We conduct ablation studies to verify the effectiveness of our RoomBench++ dataset and RoomEditor++ architecture. 

\begin{table*}[!t]
\caption{Ablation Study on RoomBench~\cite{lin2025roomeditor} vs. RoomBench++.}
\centering
\setlength{\tabcolsep}{5pt} 
\renewcommand{\arraystretch}{1.15}
\begin{tabularx}{0.9\textwidth}{c|c|*{6}{>{\centering\arraybackslash}X}}
\toprule
Testing set & Training set & FID↓ & SSIM↑ & PSNR↑ & LPIPS↓ & CLIP↑ & DINO↑ \\
\midrule
\multirow{2}{*}{RoomBench }
  & RoomBench & 18.42 & 0.793 & 21.15 & 0.094 & 90.51 & 85.47 \\
  & RoomBench++ & \textbf{15.88} & \textbf{0.862} & \textbf{22.96} & \textbf{0.085} & \textbf{91.79} & \textbf{90.42} \\
\midrule
\multirow{2}{*}{RoomBench++}
  & RoomBench & 14.78 & 0.859 & 24.75 & 0.078 & 89.43 & 83.73 \\
  & RoomBench++ & \textbf{11.93} & \textbf{0.891} & \textbf{26.10} & \textbf{0.071} & \textbf{90.99} & \textbf{87.17} \\
\bottomrule
\end{tabularx}
\label{tab:ablation_roombench++}
\end{table*}

\begin{table*}[!t]
\caption{\textbf{Ablation Study on Network Architectures.} The results are reported on RoomBench++. }
\centering
\setlength{\tabcolsep}{5pt}  
\begin{tabularx}{0.9\textwidth}{l|*{6}{>{\centering\arraybackslash}X}}
\toprule
Method & FID↓ & SSIM↑ & PSNR↑ & LPIPS↓ & CLIP↑ & DINO↑ \\
\midrule
RoomEditor++ (U) (Unshared U-Nets)  & 13.12 & 0.883 & 25.70 & 0.077 & 90.58 & 86.14 \\
RoomEditor++ (U) w/ CLIP    & \textbf{11.90} & \textbf{0.891} & \textbf{26.12} & \textbf{0.071} & 90.92 & \textbf{87.23} \\
RoomEditor++ (U)  & 11.93 & \textbf{0.891} & 26.10 & \textbf{0.071} & \textbf{90.99} & 87.17 \\
\midrule
RoomEditor++ w/ SigLIP   & 11.55 & \textbf{0.905} & \textbf{26.93} & \textbf{0.065} & 91.38 & 91.01 \\
RoomEditor++  & \textbf{11.49} & \textbf{0.905} & 26.82 & 0.067 & \textbf{91.39} & \textbf{91.03} \\
\bottomrule
\end{tabularx}
\label{tab:ablation_study_indoor_bench}
\end{table*}
\subsubsection{\textbf{Ablation on RoomBench++ Dataset}}

We evaluate RoomEditor++ (U) on the datasets RoomBench++ and RoomBench, where RoomBench is identically same with the realistic subset of RoomBench++.  \autoref{tab:ablation_roombench++} shows that the model trained on RoomBench++ significantly outperforms that trained on RoomBench, for both real-scene and realistic scene data. Since the models share exactly the same architecture and differ only in their training data, the performance gains substantiates the effectiveness of our RoomBench++ dataset, by introducing the large-scale real-scene subset.

\subsubsection{\textbf{Ablation on RoomEditor++ Architecture}}
Since our RoomEditor++ is highly concise, we conduct ablation experiments to evaluate its integration with key techniques (e.g., CLIP image encoder and non-shared reference U-Net $g$) in existing methods, as shown in \autoref{tab:ablation_study_indoor_bench}. First, we analyze the effects of non-shared dual U-Net configurations, where the reference U-Net $g$ is trainable. The results show that without our parameter-sharing strategy, the variant exhibits significant performance degradation. Then, we examine the integration of CLIP image encoder, which is commonly adopted to facilitate feature extraction from reference images as in MimicBrush. Although incorporating CLIP leads to a slight improvement in certain metrics, it introduces a substantial increase in the number of parameters. 

We further conduct experiments on RoomEditor++ equipped with the DiT backbone. 
First, for non-shared DiT configurations, the GPU memory cost for loading model would reach approximately 150G, which exceeds our available computational resources. Given that the results based on the U-Net backbone have already clearly demonstrated the advantages of the parameter-sharing design, these findings verify the effectiveness of our parameter-sharing diffusion architecture for unified feature learning. Thus, we omit this ablation study on the non-shared DiT architecture. 
Then, we discuss the adoption of SigLIP~\cite{zhai2023sigmoid} as an additional image encoder, a component commonly used in existing DiT-based image composition methods. We integrate SigLIP into our DiT-based RoomEditor++ and evaluate its performance. As shown in \autoref{tab:ablation_study_indoor_bench}, we come to the similar observation that introducing additional image encoder brings gains in only certain metrics. 
Considering extra computational cost, our concise dual parameter-sharing diffusion architecture is sufficient for high-fidelity furniture synthesis.  
These findings verify the effectiveness of our parameter-sharing diffusion architecture design for unified feature learning.

\section{Conclusion} \label{sec:conclusion}
This work advanced furniture synthesis by addressing two key challenges: the absence of ready-to-use benchmarks and feature space divergence in dual-branch architectures. Specifically, we collect and release RoomBench++, a ready-to-use benchmark for furniture synthesis, while presenting a parameter-sharing dual diffusion backbone (RoomEditor++) to ensure robust feature alignment. By training our RoomEditor++ on the collected RoomBench++, it achieves state-of-the-art performance for furniture synthesis in terms of geometric coherence and visual fidelity, and demonstrates strong generalization ability to diverse scenes. We hope that our work can encourage further research in furniture synthesis. 
Our method has some limitations that point to promising future directions. Since current conditional control focuses on basic mask and reference guidance, integrating physical consistency constraints (e.g., lighting changes, shadow) will enable more physically plausible furniture placement. These refinements will extend the model’s utility in practical home design scenarios.

{
    \small
    \bibliographystyle{IEEEtran}
    \bibliography{main}
}

\end{document}